\documentclass[journal]{IEEEtran}
\ifCLASSINFOpdf
\usepackage[pdftex]{graphicx}
\graphicspath{{img/}}
\DeclareGraphicsExtensions{.eps,.pdf,.tiff,.jpeg,.png}
\else
\usepackage[dvips]{graphicx}
\graphicspath{{img/}}
\DeclareGraphicsExtensions{.eps}
\fi
\usepackage{cite,epstopdf}
\usepackage{psfrag}
\usepackage{subfigure}
\usepackage{url}
\usepackage{stfloats}
\usepackage{amsmath}
\usepackage{amsmath, bm}
\usepackage{amsmath, amssymb}
\usepackage{epsfig}
\usepackage{algorithm, algorithmic}
\usepackage{color}
\usepackage{booktabs} 
\usepackage[table]{ xcolor}
\usepackage{booktabs}
\usepackage{multirow}
\usepackage{diagbox}
\usepackage[colorlinks,linkcolor=blue]{hyperref}

\usepackage{alphalph}
\begin{document}
\title{\vspace{-2mm}CNN Injected Transformer for Image \\Exposure Correction}

\author{
		Shuning~Xu,
        ~Xiangyu~Chen,~\IEEEmembership{Student~Member,~IEEE},
        Binbin~Song,~\IEEEmembership{Student~Member,~IEEE},
        and ~Jiantao~Zhou,~\IEEEmembership{Senior Member,~IEEE}

		\vspace{-7mm}

        \thanks{
        	
        	The authors are with the State Key Laboratory of Internet of Things for Smart City and the Department of Computer and Information Science, University of Macau, Macau, China. e-mails: yc07425@umac.mo, chxy95@gmail.com, yb97426@umac.mo and jtzhou@umac.mo (\emph{Corresponding Author: Jiantao Zhou}).
         }  
         
         }

\maketitle

\begin{abstract}
Capturing images with incorrect exposure settings fails to deliver a satisfactory visual experience. Only when the exposure is properly set, can the color and details of the images be appropriately preserved.
Previous exposure correction methods based on convolutions often produce exposure deviation in images as a consequence of the restricted receptive field of convolutional kernels.
This issue arises because convolutions are not capable of capturing long-range dependencies in images accurately. To overcome this challenge, we can apply the Transformer to address the exposure correction problem, leveraging its capability in modeling long-range dependencies to capture global representation. However, solely relying on the window-based Transformer leads to visually disturbing blocking artifacts due to the application of self-attention in small patches.
In this paper, we propose a CNN Injected Transformer (CIT) to harness the individual strengths of CNN and Transformer simultaneously.
Specifically, we construct the CIT by utilizing a window-based Transformer to exploit the long-range interactions among different regions in the entire image.
Within each CIT block, we incorporate a channel attention block (CAB) and a half-instance normalization block (HINB) to assist the window-based self-attention to acquire the global statistics and refine local features.
In addition to the hybrid architecture design for exposure correction, we apply a set of carefully formulated loss functions to improve the spatial coherence and rectify potential color deviations. 
Extensive experiments demonstrate that our image exposure correction method outperforms state-of-the-art approaches in terms of both quantitative and qualitative metrics. 
The source code and pre-trained models are available at \href{https://github.com/rebeccaeexu/CIT-EC/}{https://github.com/rebeccaeexu/CIT-EC/}.

\end{abstract}

\begin{IEEEkeywords}
Image exposure correction, swin transformer, deep learning
\end{IEEEkeywords}

\IEEEpeerreviewmaketitle

\section{Introduction}
\label{sec:intro}
\begin{figure*}[htbp]
\centering
\includegraphics[width=\linewidth]{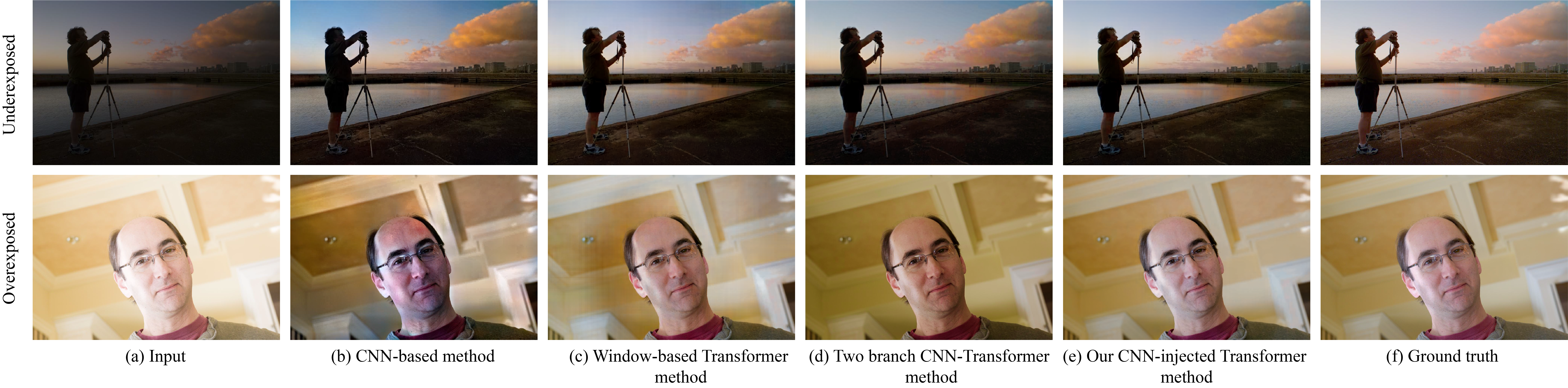}
\vspace{-3mm}
\caption{Sample photographs with underexposure and overexposure problems. (a) Input underexposed and overexposed images; Corrected results by (b) CNN-based method \cite{afifi2021learning}, (c) Transformer-based method \cite{liang2021swinir}, (d) two branch CNN-Transformer method \cite{cui2022you} and (e) our CNN-injected Transformer method; (e) Ground truth images.
\vspace{-4mm}}
\label{fig:def}
\end{figure*}

Proper exposure is crucial to achieving ideal photographs. For both conventional film cameras and digital cameras, the Exposure Value (EV) should be accurately set according to the intensity of the ambient light source. Incorrect exposure causes unsatisfactory visual problems in the captured images, including underexposure and overexposure.
Underexposure is the condition of insufficient light reaching the film or digital sensor, resulting in difficulties in discerning shadow details in the image.
As shown in the upper image of Fig. \ref{fig:def}(a), the facial features and clothing colors of the individuals are not clearly visible due to the person standing against a backlit background. Likewise, overexposure refers to an image being excessively bright, resulting in a washed-out appearance. As depicted in the lower image of Fig. \ref{fig:def}(a), the man's face appears washed out, obscuring wrinkles and whiskers, displaying a pale white tone.

Currently, photos are commonly captured using digital cameras or mobile phones in either automatic or semi-automatic exposure modes. In most instances, exposure can be estimated accurately to some extent. However, in scenes with a significant difference in lighting levels, exposure deviations may occur due to the limited dynamic range of the charge-coupled device in consumer digital cameras.
A practical approach is to correct exposure through post-processing operations, as advocated by numerous recent studies.

Earlier works developed exposure correction methods that are capable of adjusting underexposed or overexposed images separately \cite{ wang2019underexposed,srinivas2019exposure,zheng2020single,liu2022efinet,li2021low,fan2022multiscale}. When encountering a potential incorrect exposure situation, it is necessary to determine the exposure deviation, whether the image is underexposed or overexposed before applying the appropriate exposure correction model. However, the inaccurate selection of the model may further destroy the image's visual experience. Additionally, employing separate models for exposure correction introduces additional computational overhead.

Recent efforts primarily concentrated on training end-to-end deep networks using a combination of data from various exposure conditions for exposure correction \cite{afifi2021learning, huang2022exposure, huang2022deep}.	
These methods are convolution-based methods, taking the capability of CNN for extracting local features to learn the exposure correction. However, exposure correction heavily relies on utilizing global information, which is the overall illumination of the captured image.
Solely relying on CNNs poses challenges for exposure correction due to the limited receptive field of small convolutional kernels, which restricts the capture of global information. Fig. \ref{fig:def}(b) exemplifies a scenario where CNN-based methods produce deviations in overall exposure when compared to the ground truth. Furthermore, applying content-adaptive treatments is challenging when using the same convolutional kernel to restore various image regions.

As an alternative to CNN, Transformer \cite{vaswani2017attention} has shown effectiveness in various vision tasks \cite{dosovitskiy2020image} thanks to its ability of building long-range feature dependencies. 
Intuitively, Transformer could be applied to the exposure correction problem to model the global representation of the image illumination. However, self-attention is typically applied to small patches (\textit{e.g.}, 8 $\times$ 8) that are processed independently, which leads to blocking artifacts. Recently, SwinIR \cite{liang2021swinir} proposed an image restoration model to capture long-range information with the shifted window scheme. Unfortunately, when directly applying window-based Transformer to the exposure correction task, blocking artifacts persist in low-frequency regions, as evidenced by the background area in Fig. \ref{fig:def}(c). The reason for this is that the shifted window scheme does not establish enough connections among local windows.

Instinctively, coupling CNN and Transformer can complement each other's weaknesses and make it possible for us to utilize their fortes for refining localized and long-range information simultaneously. To achieve this, one simple way is to utilize a two-branch structure \cite{li2021two} with a CNN branch for the refinement of local features and a Transformer branch for the processing of global features. As demonstrated in Fig. \ref{fig:def}(d), while blocking artifacts are alleviated, the issue of color deviation in CNN-based models seems to have resurfaced. This could be attributed to the challenges associated with merging the features extracted from the two separate branches.

In this paper, we inject CNN structures directly into Transformer blocks to better utilize both the advantages of CNN and Transformer and further build stronger relations between the partitioned windows. 
Fig. \ref{fig:def}(e) demonstrates the superior performance of our CNN-injected Transformer method in minimizing exposure deviations and achieving enhanced color restoration, in comparison to the CNN-based method illustrated in  Fig. \ref{fig:def}(b). Moreover, when compared to the results obtained from the window-based Transformer method depicted in Fig. \ref{fig:def}(c), our approach effectively mitigates the unsatisfactory visual experience caused by blocking artifacts through the incorporation of CNN modules. Unlike the two-branch structure that combines Transformer and CNN, our method employs region-specific adjustments for overall exposure correction, ensuring exceptional treatment of both global and fine-grained image details. In the upper image of Fig. \ref{fig:def}(e), our method enables flexible adjustments of brightness for both the character and the background, with a stronger emphasis on character enhancement. This adjustment aligns more closely with the ground truth, as demonstrated in Fig. \ref{fig:def}(f).
We introduce the CIT block to combine Transformer and CNN. In each CIT block, we incorporate both the channel attention block (CAB) and the half-instance normalization block (HINB) using two distinct approaches. The CAB runs parallel to the Window-based multi-head self-attention (W-MSA) module, suppressing blocking artifacts. The HINB is injected parallel to the window-based Transformer block after the initial Layer Norm (LN), enabling the calibration of mean and variance of features in shallow layers to refine local image details. The employed design facilitates the correction of the overall exposure of the image while applying varying degrees of correction to specific regions within it. Furthermore, we employ a set of meticulously designed loss functions. We utilize color loss and spatial loss to respectively rectify potential color deviations and enhance spatial coherence. Extensive experimental results demonstrate the superiority of our proposed strategies over state-of-the-art methods in both quantitative and qualitative comparisons.

In summary, our contributions are listed as follows:

\begin{itemize}
\item We propose a framework combining window-based Transformer and CNN to address the exposure correction problem while avoiding the introduction of additional blocking artifacts.

\item To model long-range dependencies and enhance local details in the image, CNN-injected Transformer Blocks are devised as crucial components. We directly inject convolutional structures CAB and HINB into the transformer module, jointly considering image local and long-range information.

\item Our proposed method achieves a remarkable performance gain of 1.51dB in PSNR and 0.07 in SSIM on SICE dataset, and around 0.4dB in PSNR and 0.02 in SSIM on MSEC dataset.
\end{itemize}

The rest of the paper is organized as follows. In Section \ref{sec:relatedwork}, we review the related works about the exposure correction and vision Transformer. In Section \ref{sec:method}, we present our CNN-injected Transformer for exposure correction. In Section \ref{sec:exp}, we show the experimental comparison and ablation analysis results. Finally, the paper is concluded in Section \ref{sec:con}.

\section{Related Works}
\label{sec:relatedwork}
In this section, we briefly introduce the related works on exposure correction and vision transformer.

\subsection{Exposure Correction}
Traditional exposure correction methods utilized the image histogram to adjust the contrast of the image \cite{celik2011contextual, lee2013contrast, reza2004realization, thomas2011histogram}. Later, some works relied on Retinex theory to estimate illumination effectively while preserving naturalness of the image \cite{guo2016lime, zhang2018high, meylan2006high, cai2017joint}. More recent efforts used deep learning for correcting exposure errors, e.g., the CNN-based approaches \cite{guo2016lime2, fu2016fusion , nsamp2021learning, huang2019hybrid, jiang2021enlightengan,liu2021retinex, huang2022low}. These methods were restricted to address one of the exposure errors, namely, addressing either the overexposure or the underexposure problem, causing poor generalization to various exposures. In contrast, several deep learning methods explicitly focused on exposure correction for both underexposed and overexposed images \cite{afifi2021learning, huang2022exposure, huang2022deep, eyiokur2022exposure, cui2022you, wang2023decoupling, huang2023learning}. Among these methods, Afifi \textit{et al.} \cite{afifi2021learning} proposed a coarse-to-fine deep network to progressively correct exposure errors. 
Huang \textit{et al.} \cite{huang2022exposure} designed a multiple exposure correction framework based on an exposure normalization and compensation module. 
Huang \textit{et al.} \cite{huang2022deep} presented a spatial-frequency interaction network to progressively reconstruct the representation of lightness and structure components. 
Cui \textit{et al.} designed a two-branch Transformer model, preserving pixel-wise informative details in the local branch and estimating ISP components in the global ISP separately \cite{cui2022you}.
Wang \textit{et al.} \cite{wang2023decoupling} decoupled the contrast enhancement and detail restoration within each convolution process.
Huang \textit{et al.} \cite{huang2023learning} explored the relationship between underexposure and overexposure samples to solve the optimization inconsistency problem of exposure correction.
Although these data-driven approaches performed better than traditional solutions, there still exist artifacts for scenes with various exposure values.

\subsection{Vision Transformer}
Transformer was originally designed to solve sequence-to-sequence tasks in natural language processing \cite{vaswani2017attention}. In computer vision, attention mechanism of Transformer can also be applied to the sequences of non-overlapping image patches, and shows impressive performance in high-level vision tasks \cite{vaswani2021scaling, carion2020end, chu2021twins, wang2021pyramid}.
Despite the proven effectiveness of vision Transformers in modeling long-range dependencies, numerous studies have demonstrated that convolutions can enhance the ability of Transformers to achieve superior visual representations in various low-level vision tasks \cite{chen2021pre, cao2021video, liang2021swinir, zamir2022restormer, chen2023activating, wang2022uformer}.
SwinIR \cite{liang2021swinir} proposed an image restoration model based on Swin Transformer \cite{liu2021swin}, demonstrating superior performance in both image super-resolution and denoising tasks.
Uformer \cite{wang2022uformer} introduced a hierarchical encoder-decoder network using the Transformer block for image restoration task.
Restormer \cite{zamir2022restormer} improved feature aggregation and transformation to learn long-range dependencies while remaining computationally efficient.
Our method is partially motivated by these transformer-based methods. 
In this paper, we propose a novel approach for the exposure correction task by combining CNN with Transformer-based models, aiming to leverage the benefits of both CNN and Transformer.
 
\section{CNN Injected Transformer for Exposure Correction}
\label{sec:method}

In this section, we first formulate the process of exposure correction in Section \ref{sec:motivation}. In Section \ref{sec:CITB}, a CNN Injected Transformer is proposed to learn the global and local exposure representation of the image. Finally, the training strategy and objectives are presented in Section \ref{sec:loss}.

\subsection{Motivation and Overview}
\label{sec:motivation}

\begin{figure*}[t]
\centering
\includegraphics[width=\linewidth]{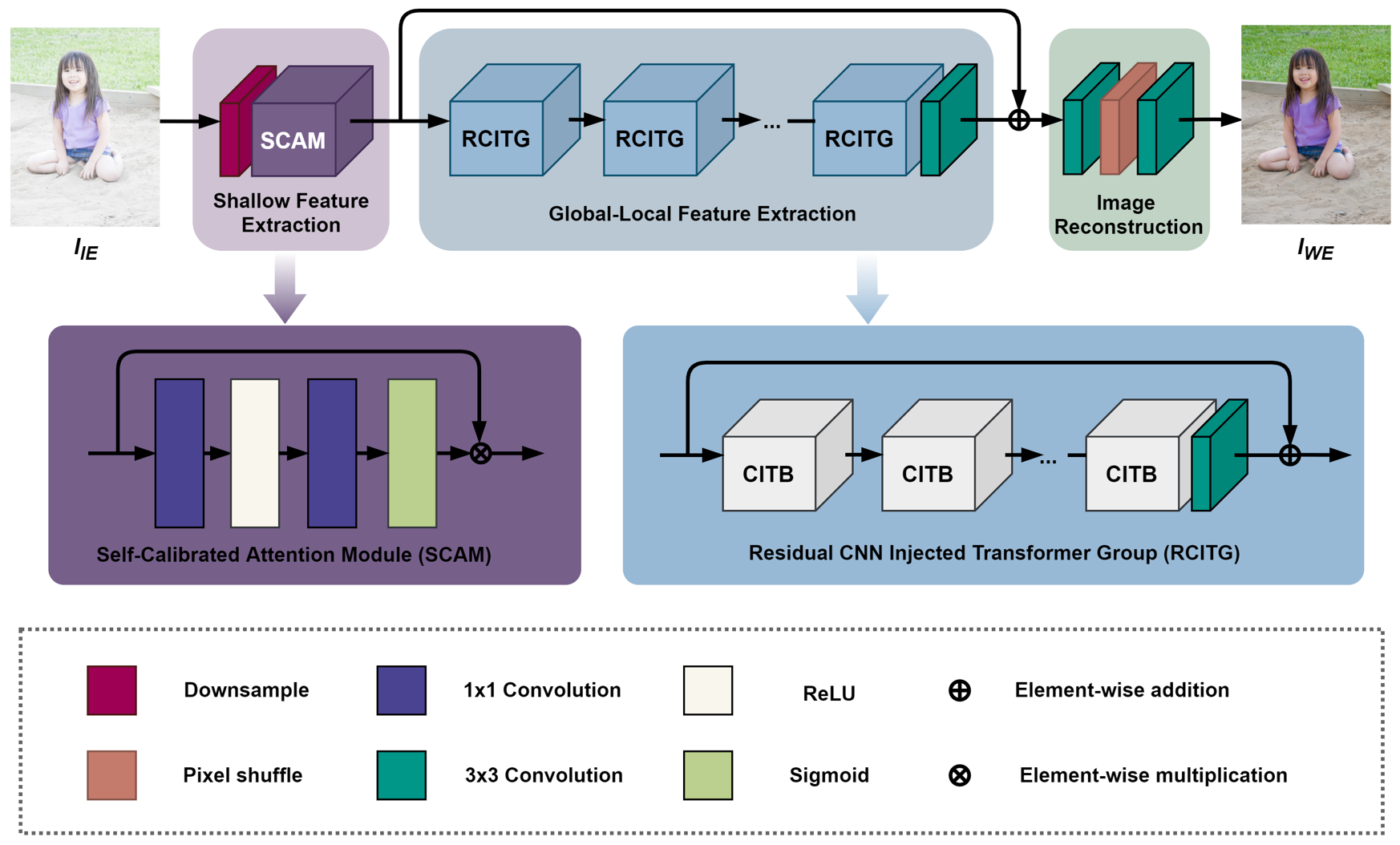}
\vspace{-6mm}
\caption{Overview of our CNN Injected Transformer architecture for exposure correction. Global-Local Feature Extraction stage consists of several RCITGs (Residual CNN Injected Transformer Groups). Each RCITG is composed of several CITBs (CNN Injected Transformer Blocks). \vspace{-4mm}}
\label{fig:main}
\end{figure*}

Given an incorrectly-exposed input image $I_{IE}$, which can be underexposed or overexposed, the aim of exposure correction is to restore it to a well-exposed image $I_{WE}$. As depicted in Fig. \ref{fig:main}, the overall framework of our proposed method comprises shallow feature extraction, global-local feature extraction and image reconstruction.

For $I_{IE}\in\mathbb{R}^{H\times W\times 3}$, where $H$ and $W$ are height and width of the input image, we first apply shallow feature extraction $H_{SF}$ to extract the shallow features $F_0\in\mathbb{R}^{\frac{H}{4}\times \frac{W}{4}\times C}$ as:
\begin{equation}
   F_0=H_{SF}(I_{IE}),
   \vspace{-1.5mm}
\end{equation}
where $C$ is the channel number of intermediate features. The size of intermediate features is down-sampled to $\frac{H}{4} \times \frac{W}{4}$.
$H_{SF}$ comprises a single convolutional layer with a stride of four and the self-calibrated attention module $H_{SCAM}$. $H_{SCAM}$ can be viewed as a pixel attention mechanism, constructed by combining two 1$\times$1 convolutional layers, an activation function, and a sigmoid function. Incorporating 1$\times$1 convolutions allows for modeling non-linear transformations between channels of feature maps, enabling the capture of complex inter-channel relationships. This enhances the network's representational power and its ability to model complex patterns in incorrectly-exposed image features across diverse scenes, thereby preserving the original image details during exposure correction.

After obtaining the down-sampled shallow features $F_0$, we employ global-local feature extraction $H_{GLF}$ to learn the global and local exposure features $F_{GLF}\in\mathbb{R}^{\frac{H}{4}\times \frac{W}{4}\times C}$ as:
\begin{equation}
\vspace{-1.5mm}
   F_{GLF}=H_{GLF}(F_0),
\end{equation}
where $H_{GLF}$ consists of $N$ Residual CNN Injected Transformer groups (RCITG) and one 3 $\times$ 3 convolutional layer $H_{CONV}$.
The core part of RCITG is the CNN injected Transformer block (CITB). In each window-based Transformer block, we incorporate two distinct CNN-based structures, leveraging the Transformer's capability to capture long-range dependencies and the CNN's proficiency in refining the most informative local features. A comprehensive description of CITB can be found in Section \ref{sec:CITB}.
Subsequently, we incorporate a global residual connection to integrate the shallow and global-local features. Finally, we employ the image reconstruction module $H_{REC}$ to derive $I_{WE}\in\mathbb{R}^{H\times W\times 3}$ as:
\begin{equation}
   I_{WE}=H_{REC}(F_0+F_{GLF}),
\end{equation}
where $H_{REC}$ consists of several convolutional layers and a sub-pixel convolution layer to upsample the features to reconstruct the well-exposed image with height and width of $H$ and $W$.

\subsection{CNN Injected Transformer}
\label{sec:CITB}
Exposure serves as a comprehensive representation of an image, encompassing its global characteristics. Nevertheless, in scenes with significant lighting variations, the bright areas exhibit distinct illuminance characteristics compared to the shadow regions. Employing solely convolutional-based modules for exposure correction hinders the acquisition of global information, while utilizing a uniform convolutional kernel for restoring diverse illumination regions poses challenges in applying context-specific adjustments. Transformers possess the capability to capture long-range dependencies, a crucial aspect required for exposure correction. Nevertheless, treating each pixel as a separate token in Transformer architectures incurs a significant computational cost. Window-based Transformers \cite{liang2021swinir} are widely adopted in the field of image restoration due to their ability to effectively reduce the computational burden. Regrettably, the adoption of window-based Transformers introduces new challenges. The window-partition operation in such Transformers can introduce blocking artifacts into the restored image, significantly compromising the image quality subsequent to exposure correction.

Consequently, we present to utilize the convolutional network to mitigate the blocking artifacts introduced by the window-based transformer.
Based on this idea, we introduce the CITB, which constitutes the $H_{GLF}$. $H_{GLF}$ encompasses multiple RCITG and a convolutional layer, where each RCITG comprises several CITB and a convolutional layer. Now we explain the structures of RCITG and CITB in detail.

\subsubsection{Residual CNN Injected Transformer Group}
As shown in Fig. \ref{fig:main}, we propose the global-local feature extraction $H_{GLF}$ to learn the global and local exposure representation of the image. $H_{GLF}$ consists of $N$ RCITG, and one 3 $\times$ 3 convolutional layer $H_{CONV}$. In particular, intermediate features $F_i$ and the output feature $F_{GLF}$ are progressively extracted as:
\begin{gather}
\label{equ:RCITG}
\vspace{-1.5mm}
  F_{i}=H_{RCITG_i}(F_{i-1}), i=1,2,...,N, \notag \\ 
  F_{GLF}=H_{CONV}(F_N), 
\end{gather}
where $H_{RCITG_i}$ represents the $i$-th RCITG. Following \cite{liang2021swinir}, we apply a convolutional layer at the end of each RCITG to better aggregate the features.

Each RCITG consists of $M$ CITBs and one $3\times 3$ convolutional layer with a residual structure. 
Given the input feature $F_{i,0}$ of the $i$-th RCITG, the intermediate features $F_{i,1}$, $F_{i,2}$, \ldots, $F_{i,M}$ are extracted as:
\begin{equation}
  F_{i,j}=H_{CITB_{i,j}}(F_{i,j-1}), \quad j=1,2,...,M, \\ 
\end{equation}
where $H_{CITB_{i,j}}$ is the $j$-th CITB in the $i$-th RCITG.
After the mapping of a series of CITBs, we reserve the convolutional layer $H_{CONV_i}$ before the residual connection in $i$-th RCITG following \cite{liang2021swinir} as:
\begin{equation}
\vspace{-1.5mm}
  F_{i}=H_{CONV_i}(F_{i,M})+F_{i-1,0}.
\end{equation}
A residual connection is added for better convergence in training process.

\begin{figure}[t]
\vspace{-0.3cm}
\centering
\includegraphics[width=\linewidth]{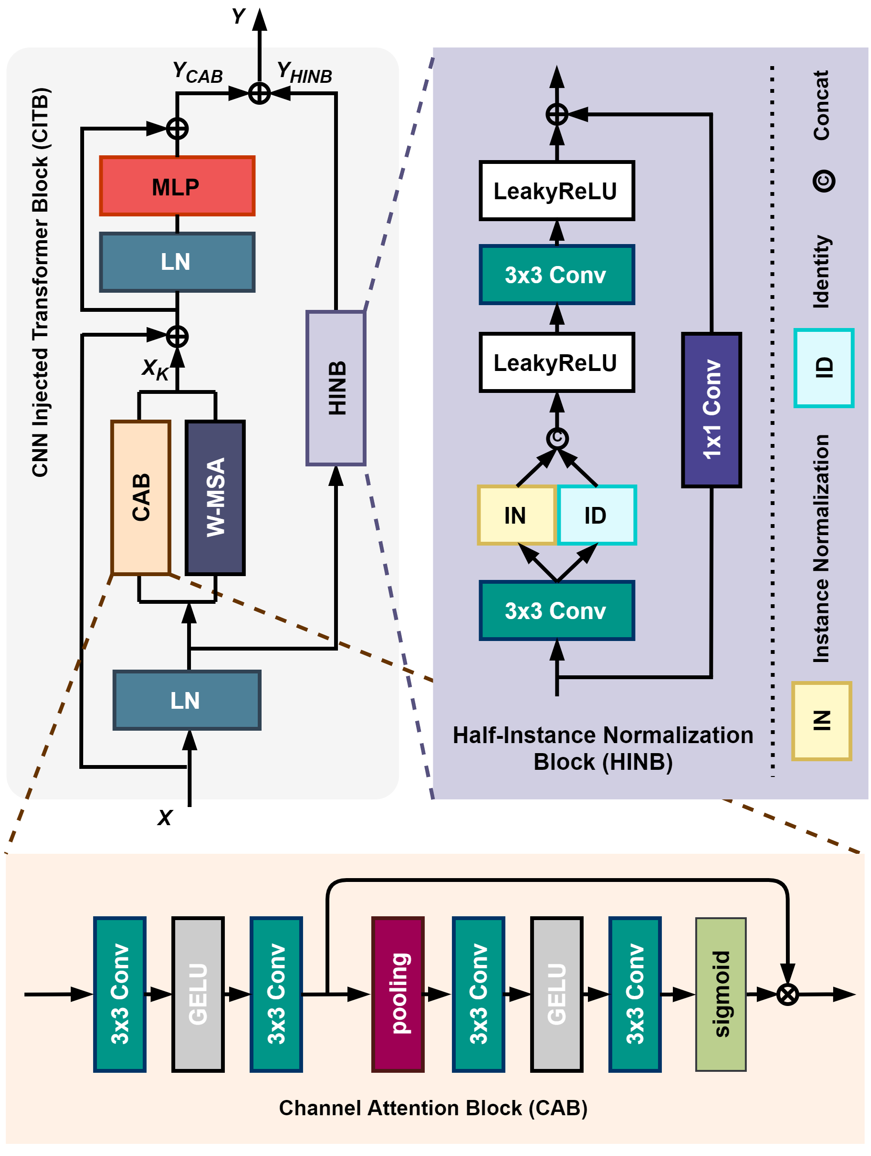}
\vspace{-6mm}
\caption{The structure of CNN Injected Transformer Block (CITB).}
\vspace{-6mm}
\label{fig:cit}
\end{figure}

\subsubsection{CNN Injected Transformer Block} 
Modifying the exposure values directly affects the overall illumination of the captured image. To obtain a comprehensive representation of the image, we select a window-based Transformer structure as the fundamental component of the $H_{GLF}$ module, enabling the capture of global information. The local correlation between pixel values of neighboring pixels in an image is significantly impacted by the window-partition mechanism of the conventional window-based Transformer \cite{liu2021swin}, leading to impairment. However, a standalone window-based Transformer is inadequate for capturing the local correlation between windows, resulting in the presence of blocking artifacts in the final reconstructed well-exposed images. Consequently, we incorporate CNN blocks directly into each window-based Transformer, forming the CITB.
As shown in in Fig. \ref{fig:cit}, inspired by \cite{chen2023activating}, we incorporate a channel attention block (CAB) into the window-based Transformer, and further introduce a half-instance normalization block (HINB) as well. This integration enables us to effectively leverage the strengths of both CNN and Transformer to refine the most informative local features and model long-range dependencies.

This injection process can be formulated as:
\begin{equation}
\vspace{-1.5mm}
  Y=Y_{CAB}+Y_{HINB},
\end{equation}
where $Y_{CAB}$, $Y_{HINB}$ and $Y$ denote the output of CAB, HINB and CITB, respectively.

In parallel with the window-based multi-head self-attention module (W-MSA), a CAB is inserted, and it is followed by layer normalization (LN). In various image restoration tasks, introducing CAB results in the activation of more pixels due to the inclusion of global information in the calculation of channel attention weights \cite{chen2023activating,song2022multistage,song2023real}. As for exposure correction, the model can refine local information by reshaping flattened token sequences back to the spatial grid, thereby alleviating blocking artifacts caused by the window-partition mechanism. Additionally, this operation facilitates the restoration of image details by selecting cross-channel features.
It should be noted that, similar to \cite{liu2021swin, liang2021swinir}, we utilize shifted window-based multi-head self-attention (SW-MSA) at intervals within consecutive CITBs.
For a local window feature $X \in \mathbb{R}^{W^2 \times C}$, the CAB-injected branch in CITB can be formulated as:
\begin{gather}
  X_K=MSA(LN(X))+\alpha CAB(LN(X))+X, \notag \\
  Y_{CAB}=MLP(LN(X_K))+X_K,
\end{gather}
where $W\times W$ is the size of the non-overlapping local windows and $X_K$ denotes the intermediate features in CAB-injected branch. For easier optimization and better visual effect, a small constant $\alpha$ is multiplied to the output of CAB.

To refine local image details and facilitate overall exposure correction while allowing varying degrees of correction in specific regions, the HINB is injected in parallel with the Transformer block after applying LN.
Instance normalization (IN) can recalibrate the mean and variance of features without affecting the batch dimension. HINBs utilize IN on half of the channels while preserving the context information as identity (ID) in the other half of the channels \cite{chen2021hinet}. This approach enhances the effectiveness and efficiency of calibrating the mean and variance of features, particularly in shallow layers.
The process of HINB-branch in CITB is computed as:
\begin{equation}
    Y_{HINB} = \beta HINB(LN(X)),
\end{equation}
where $\beta$ is the small constant.

\subsection{Training Strategy and Objectives}
\label{sec:loss}
We train our framework in an end-to-end manner and the overall training objective can be expressed as:
\begin{equation}
\label{eq:final_loss}
{L} = {L}_{rec} + \lambda_1 {L}_{col} + \lambda_2 {L}_{spa},
\end{equation}
where \textit{$L_{rec}$} is the reconstruction loss, \textit{$L_{col}$} is the color loss, and \textit{$L_{spa}$} is the spatial loss. Color loss $L_{col}$ and spatial loss $L_{spa}$ are inherited from Zero-DCE \cite{guo2020zero}, as the exposure correction task shares some similar targets with low-light image enhancement. We empirically set the weights $\lambda_1=0.5$ and $\lambda_2=0.5$. Details are given below.

\textbf{Reconstruction loss.} We use $L1$ loss function between the reconstructed well-exposed images $I_{WE}$ and the ground truth images $I_{GT}$ as:
\begin{equation}
\label{eq:rec_loss}
{L}_{rec} = \|I_{GT} - I_{WE}\|_1.
\end{equation}

\textbf{Color loss.} 
Gray-World color constancy hypothesis ~\cite{buchsbaum1980spatial} illustrates that color in each sensor channel averages to gray over the entire image. To correct the potential color deviations and build the relations among the three channels, we apply color loss $L_{col}$ as:
\begin{equation}
\label{equ:color_loss}
L_{col}=\sum\nolimits_{\forall(p,q)\in \mathcal{S}}(I_{WE}^{p}-I_{WE}^{q})^2, \mathcal{S}
=\{(r,g),(g,b),(g,b)\},
\end{equation}
where $I_{WE}^{p}$ and $I_{WE}^{q}$ denote the average intensity value of the $p$ and $q$ color channels in the enhanced image respectively.
The color channel pair $(p, q)$ is a set consisting of all possible combinations formed by selecting two channels from the three color channels. This set $S$ includes all such pairs, namely $(r,g),(g,b),(g,b)$.

\textbf{Spatial loss.} To maintain the spatial coherence of the exposure corrected image, we minimize the difference in neighbouring regions between the incorrectly exposed input images $I_{IE}$ and the well-exposed output images $I_{WE}$ as:
\begin{equation}
\label{equ:spa_loss}
L_{spa}=\frac{1}{K}\sum\limits_{k=1}^K(Mean(I_{WE})-Mean(I_{IE}))^2,
\end{equation}
where $K$ is the number of non-overlapping local regions of size 4 $\times$ 4. $Mean(\cdot)$ denotes the average intensity values of $k$-th local region.

\section{Experiments}
\label{sec:exp}

\subsection{Experimental Setup}
\subsubsection{Datasets}
The experiments are conducted on the dataset collected by MSEC \cite{afifi2021learning}, which is rendered from the MIT-Adobe FiveK dataset \cite{bychkovsky2011learning}. This dataset provides 24,330 8 bit-sRGB images with various exposure settings and their corresponding expert-retouched ground truth images. Specifically, it is separated into 3 sets: (\romannumeral1) 17675 image pairs for training, (\romannumeral2) 750 image pairs for validation, (\romannumeral3) 5905 image pairs for testing. 
Each improperly exposed image is rendered with different EVs (-1.5 EV, -1 EV, 0 EV, 1 EV, 1.5 EV).
As for the ground truth images in MSEC \cite{afifi2021learning}, five expert photographers (Expert A, B, C, D, E) are invited to do retouching manually on the improperly exposed-images. Although there are different white balancing or tone-mapping preferences for different experts, they are all able to correct exposure errors to a large extent. Rather than considering the images retouched by a single expert as the ground truth, the evaluation is conducted on the enhanced images from all the five experts.

\subsubsection{Comparative Methods} To verify the effectiveness of our method, we perform quantitative comparisons with the following methods, including both non-deep learning and learning-based solutions. Specifically, we compare with HE \cite{gonzales2001digital}, CLANE \cite{adaptivehisteq}, WVM \cite{fu2016weighted}, LIME \cite{guo2016lime, guo2016lime2}, HDRCNN \cite{HDRCNN}, DPED \cite{DPED}, DPE \cite{DPE}, HQEC \cite{HQEC}, RetinexNet \cite{Chen2018Retinex}, Deep UPE \cite{DeepUPE}, Zero-DCE \cite{guo2020zero}, MSEC \cite{afifi2021learning}, ENC \cite{huang2022exposure}, FECNet \cite{huang2022deep} and IAT \cite{cui2022you}. 
Among them, HE, CLANE, WVM, LIME and HQEC are non-deep learning methods. LIME, HQEC, RetinexNet, Deep UPE and Zero-DCE are originally designed for low-light enhancement. For HDRCNN, the reconstructed HDR images need to be converted to LDR space using HDR transformation method (RHT) or Adobe Photoshop's (PS) HDR tool, namely HDRCNN w/ RHT and HDRCNN w/ PS.

\subsubsection{Implementation Details} We train our model on 4 NVIDIA Tesla A100 GPUs and conduct the evaluation on 1 NVIDIA 2080 Ti GPU. For the structure of our proposed model, the RCITG number, CITB number, window size, channel number and attention head number are set to 6, 6, 8, 180 and 6, respectively. For the hyper-parameter in CITB, we set the weighting factors of CAB output ($\alpha$) and HINB output ($\beta$), and the squeeze factor between two convolution layers as 0.01, 0.01 and 3, respectively.
During the training stage, each image is randomly cropped into patches of size 256 $\times$ 256 as inputs.
We use the Adam optimizer and set the batch size as 32 and fix learning rate as 1 $\times$ $10^{-4}$, respectively. The network is trained for 15 epochs and the training process costs about 12 hours. 

\subsubsection{Evaluation Criteria}
We adopt the following three standard metrics to evaluate the pixel-wise accuracy and the perceptual quality of our results:
Peak Signal-to-Noise Ratio (PSNR), Structural Similarity Index Measure (SSIM) \cite{wang2004image}, and perceptual index (PI) \cite{blau20182018} as evaluation metrics. The $PI$ is formed with:
\begin{equation}
\label{eq:PI}
    PI = 0.5(10 - Ma + NIQE),
\end{equation}
where both $Ma$ \cite{ma2017learning} and $NIQE$ \cite{mittal2012making} are no-reference image quality metrics.

\subsection{Quantitative Results}
Table \ref{table:results1} shows the quantitative comparison on MSEC testing set, which is divided into three parts: (\romannumeral1) properly exposed and over-exposed images, (\romannumeral2) under-exposed images, (\romannumeral3) combined over- and under-exposed images. 

Deep learning-based exposure correction methods, including MSEC, ENC, FECNet, and IAT, surpass the traditional methods and those tailored for low-light scenarios. As for restoring low-light images, our approach achieves exceptional performance, as shown in the second part in Table \ref{table:results1}.

We discover that the IAT method, which utilizes a two-branch structure combining CNN and Transformer, cannot achieve comparable numerical performance when compared to meticulously designed CNN-based methods such as ENC and FECNet. This discrepancy may be attributed to the two-branch design of IAT, which struggles to effectively integrate global information and local information of the image. In contrast, our CNN-injected Transformer approach exhibits remarkable superiority over both the IAT method and other CNN-based methods in terms of numerical metrics. Notably, our model outperforms the second-best method by 0.4dB in PSNR and 0.01 in SSIM on the exposure-combined MSEC dataset. Regarding the no-reference image quality metric PI, our method demonstrates a commendable score, trailing only marginally behind the leading method by a mere 0.087.

\begin{table*}
\caption{Quantitative results on MSEC \cite{afifi2021learning} testing set. \textbf{The best results are highlighted with bold.} \underline{The second-best results are highlighted } \underline{with underline.} We compare each method with properly exposed reference image sets rendered by five expert photographers.  For each method, we present peak signal-to-noise ratio (PSNR), structural similarity index measure (SSIM) and perceptual index (PI). Non-deep learning methods are marked by $\textasteriskcentered$.
\vspace{-5mm}}
\label{table:results1}
\begin{center}
\scalebox{0.92}{
\begin{tabular}{|l|c|c|c|c|c|c|c|c|c|c|c|c|c|}
\cline{1-14}
\multirow{2}{*}{Method} & \multicolumn{2}{c|}{Expert A} & \multicolumn{2}{c|}{Expert B} & \multicolumn{2}{c|}{Expert C} & \multicolumn{2}{c|}{Expert D} & \multicolumn{2}{c|}{Expert E} & \multicolumn{2}{c|}{Avg.} & \multirow{2}{*}{PI} \\ \cline{2-13} 
 & PSNR & SSIM & PSNR & SSIM & PSNR & SSIM & PSNR & SSIM & PSNR & SSIM & PSNR & SSIM & \\ \cline{1-14} \hline

\multicolumn{14}{|c|}{\cellcolor[HTML]{CCECEB}$+0$, $+1$, and $+1.5$ relative EVs (3,543 properly exposed and overexposed images)}\\ \hline
HE \cite{gonzales2001digital} $\textasteriskcentered$& 16.140 & 0.686 & 16.277 & 0.672 & 16.531 & 0.699 & 16.643 & 0.669 & 17.321 & 0.691 & 16.582 & 0.683 & 2.351\\
CLAHE \cite{adaptivehisteq} $\textasteriskcentered$& 13.934 & 0.568 & 14.689 & 0.586 & 14.453 &  0.584 & 15.116 & 0.593 & 15.850 & 0.612 & 14.808 & 0.589 & 2.270\\
WVM \cite{fu2016weighted} $\textasteriskcentered$& 12.355 & 0.624 & 13.147 & 0.656 & 12.748 & 0.645 & 14.059 & 0.669 & 15.207 & 0.690 & 13.503 & 0.657 & 2.342\\
LIME \cite{guo2016lime, guo2016lime2} $\textasteriskcentered$& 9.627 & 0.549 & 10.096 & 0.569 & 9.875 & 0.570 & 10.936 & 0.597 & 11.903 & 0.626 & 10.487 & 0.582 & 2.412\\
HDRCNN \cite{HDRCNN} w/ RHT \cite{yang2018image}& 13.151 & 0.475 & 13.637 & 0.478 & 13.622 & 0.497 & 14.177 & 0.479 & 14.625 & 0.503 & 13.842 & 0.486 & 4.284\\
HDRCNN \cite{HDRCNN} w/ PS \cite{dayley2010photoshop}& 14.804  & 0.651 & 15.622 & 0.689 & 15.348 & 0.670 & 16.583  & 0.685 & 18.022 & 0.703 & 16.076 & 0.680 & 2.248\\
DPED (iPhone) \cite{DPED}& 12.680  &  0.562 & 13.422 & 0.586 & 13.135 & 0.581 & 14.477 & 0.596 & 15.702 & 0.630 & 13.883 & 0.591 & 2.909\\
DPED (BlackBerry) \cite{DPED} & 15.170 & 0.621 & 16.193 & 0.691 &  15.781 & 0.642 & 17.042 & 0.677 & 18.035 & 0.678 & 16.444 & 0.662 & 2.518 \\
DPED (Sony) \cite{DPED}& 16.398 & 0.672 & 17.679 & 0.707 & 17.378 & 0.697 & 17.997 & 0.685 & 18.685 & 0.700 & 17.627 & 0.692 & 2.740\\
DPE (HDR) \cite{DPE} & 14.399 & 0.572 & 15.219 & 0.573 & 15.091 & 0.593 & 15.692 & 0.581 & 16.640 & 0.626 & 15.408 & 0.589 & 2.417 \\
DPE (U-FiveK) \cite{DPE} & 14.314 & 0.615 & 14.958 & 0.628 & 15.075 & 0.645 & 15.987 & 0.647 & 16.931 & 0.667 & 15.453 & 0.640 & 2.630\\
DPE (S-FiveK) \cite{DPE} & 14.786 & 0.638 & 15.519 & 0.649 & 15.625 &  0.668 & 16.586 & 0.664 &  17.661 & 0.684 & 16.035 & 0.661 & 2.621\\
HQEC \cite{HQEC} $\textasteriskcentered$& 11.775 & 0.607 & 12.536 & 0.631 & 12.127 & 0.627 & 13.424 & 0.652 & 14.511 & 0.675 & 12.875 & 0.638 & 2.387\\
RetinexNet \cite{Chen2018Retinex} & 10.149 & 0.570 & 10.880 & 0.586 & 10.471 & 0.595 & 11.498 & 0.613 & 12.295 & 0.635 & 11.059 & 0.600 & 2.933\\
Deep UPE \cite{DeepUPE} & 10.047 & 0.532 & 10.462 & 0.568 & 10.307 & 0.557 & 11.583 & 0.591 & 12.639 & 0.619 & 11.008 & 0.573 & 2.428\\
Zero-DCE \cite{guo2020zero}  & 10.116 & 0.503 & 10.767 & 0.502 & 10.395 & 0.514 & 11.471 & 0.522 & 12.354 & 0.557 & 11.021 & 0.520 & 2.774\\
MSEC \cite{afifi2021learning} & 18.976 & 0.743 & 19.767 & 0.731 & 19.980 & 0.768 & 18.966 & 0.716 & 19.056 & 0.727 & 19.349 & 0.737 & \underline{2.189}\\
ENC \cite{huang2022exposure} & 19.820 & 0.834 & 22.400 & \textbf{0.903} & 22.110 & 0.852 & \underline{21.173} & \underline{0.875} & \textbf{21.950} & \underline{0.888} & 21.490 & \underline{0.870} & \textbf{2.162}\\
FECNet \cite{huang2022deep} & \textbf{20.896} & \underline{0.837} & \textbf{23.008} & \underline{0.873} & \textbf{23.149} & \underline{0.881} & 20.823 & 0.854 & 20.336 & 0.852 & \underline{21.642} & 0.859 & 2.337 \\
IAT \cite{cui2022you} & 19.511 & 0.810 & 20.962 & 0.846 & 21.196 & 0.852 & 19.798 & 0.835 & 19.754 & 0.835 & 20.244 & 0.836 & 2.671\\
Ours & \underline{20.482} & \textbf{0.859} & \underline{22.748} & \textbf{0.903} & \underline{23.015} & \textbf{0.905} & \textbf{21.478} & \textbf{0.891} & \underline{21.679} & \textbf{0.895} & \textbf{21.881} & \textbf{0.891} & 2.293\\\hline

\multicolumn{14}{|c|}{\cellcolor[HTML]{CCECEB}$-1$ and $-1.5$ relative EVs (2,362 underexposed images)}\\ \hline
HE \cite{gonzales2001digital} $\textasteriskcentered$& 16.158 & 0.683 & 16.293 & 0.669 & 16.517 & 0.692 & 16.632 & 0.665 & 17.280 & 0.684 & 16.576 & 0.679 & 2.486 \\
CLAHE \cite{adaptivehisteq} $\textasteriskcentered$& 16.310 & 0.619 & 17.140 & 0.646 & 16.779 & 0.621 & 15.955 & 0.613 & 15.568  & 0.608 & 16.350 & 0.621 & 2.387\\
WVM \cite{fu2016weighted} $\textasteriskcentered$& 17.686 & 0.728 & 19.787 & 0.764 & 18.670 & 0.728 & 18.568 & 0.729 & 18.362 & 0.724 & 18.615 & 0.735 & 2.525\\
LIME \cite{guo2016lime, guo2016lime2} $\textasteriskcentered$& 13.444 & 0.653 & 14.426 & 0.672 & 13.980 & 0.663 & 15.190 & 0.673 & 16.177 & 0.694 & 14.643 & 0.671 & 2.462\\
HDRCNN \cite{HDRCNN} w/ RHT \cite{yang2018image}& 14.547 & 0.456 & 14.347 & 0.427 & 14.068 & 0.441 & 13.025 & 0.398 &  11.957 & 0.379 & 13.589 & 0.420 & 5.072\\
HDRCNN \cite{HDRCNN} w/ PS \cite{dayley2010photoshop}& 17.324 & 0.692 & 18.992 & 0.714 & 18.047 & 0.696 & 18.377 & 0.689 & 19.593 & 0.701 & 18.467 & 0.698 & 2.294\\
DPED (iPhone) \cite{DPED}& 18.814 & 0.680 & 21.129 & 0.712 & 20.064 & 0.683 &  19.711  & 0.675 & 19.574  & 0.676 & 19.858 & 0.685 & 2.894\\
DPED (BlackBerry) \cite{DPED} & 19.519 & 0.673 & 22.333 & 0.745 & 20.342 & 0.669 & 19.611 & 0.683 & 18.489 & 0.653 & 20.059 & 0.685 & 2.633 \\
DPED (Sony) \cite{DPED}& 18.952 & 0.679 & 20.072 & 0.691 & 18.982 &  0.662 & 17.450 & 0.629 & 15.857 & 0.601 & 18.263 & 0.652 & 2.905\\
DPE (HDR) \cite{DPE} & 17.625 & 0.675 & 18.542 & 0.705 & 18.127  & 0.677 & 16.831 & 0.665 & 15.891 & 0.643 & 17.403 & 0.673 & 2.340\\
DPE (U-FiveK) \cite{DPE} & 19.130 & 0.709 & 19.574 & 0.674 & 19.479 & 0.711 & 17.924 & 0.665 & 16.370 & 0.625 & 18.495 & 0.677 & 2.571\\
DPE (S-FiveK) \cite{DPE} & 20.153 & 0.738 & 20.973 & 0.697 & 20.915 & 0.738 & 19.050 & 0.688 & 17.510 & 0.648 & 19.720 & 0.702 & 2.564\\
HQEC \cite{HQEC} $\textasteriskcentered$& 15.801 & 0.692 & 17.371 & 0.718 & 16.587 & 0.700 & 17.090 & 0.705 & 17.675 & 0.716 & 16.905 & 0.706 & 2.532\\
RetinexNet \cite{Chen2018Retinex}  & 11.676 & 0.607 & 12.711 & 0.611 & 12.132 & 0.621 & 12.720 & 0.618 & 13.233 & 0.637 & 12.494 & 0.619 & 3.362\\
Deep UPE \cite{DeepUPE} & 17.832 & 0.728 & 19.059 & 0.754 & 18.763 & 0.745 & 19.641 & 0.737 & \underline{20.237} & 0.740 & 19.106 & 0.741 & 2.371\\ 
Zero-DCE \cite{guo2020zero}  & 13.935 & 0.585 & 15.239 & 0.593 & 14.552 & 0.589 & 15.202 & 0.587 & 15.893 & 0.614 & 14.964 & 0.594 & 3.001 \\
MSEC \cite{afifi2021learning} & 19.432 & 0.750 & 20.590 & 0.739 & 20.542 & 0.770 & 18.989 & 0.723 & 18.874 & 0.727 & 19.685 & 0.742 &2.344 \\
ENC \cite{cui2022you} & 20.680 & \underline{0.842} & \underline{22.940} & \underline{0.886} & \underline{22.720} & 0.854 & \underline{20.449} & \underline{0.861} & 19.71 & \underline{0.847} & \underline{21.300} & \underline{0.858} & \textbf{2.308}\\
FECNet \cite{huang2022deep} & \underline{20.812} & 0.829 & 22.392 & 0.857 & 22.521 & \underline{0.865} & 19.730 & 0.834 & 18.488 & 0.815 & 20.788 & 0.840 & 2.486\\
IAT \cite{cui2022you} & 19.784 & 0.783 & 21.649 & 0.814 & 21.374 & 0.818 & 19.663 & 0.809 & 19.157 & 0.797 & 20.325 & 0.804 & 2.410\\
Ours & \textbf{20.864} & \textbf{0.861} & \textbf{23.014} & \textbf{0.891} & \textbf{23.331} & \textbf{0.901} & \textbf{20.988} & \textbf{0.880} & \textbf{20.246} & \textbf{0.867} & \textbf{21.688} & \textbf{0.880} & \underline{2.327} \\\hline

\multicolumn{14}{|c|}{\cellcolor[HTML]{CCECEB}Combined over and underexposed images (5,905 images)}\\ \hline
HE \cite{gonzales2001digital} $\textasteriskcentered$& 16.148 & 0.685 & 16.283 & 0.671 & 16.525 & 0.696 & 16.639 & 0.668 & 17.305 & 0.688 & 16.580 & 0.682 & 2.405 \\
CLAHE \cite{adaptivehisteq} $\textasteriskcentered$& 14.884 & 0.589 & 15.669 &  0.610 & 15.383 & 0.599 & 15.452 & 0.601 & 15.737 & 0.610 & 15.425 & 0.602 & 2.317\\
WVM \cite{fu2016weighted} $\textasteriskcentered$&  14.488 & 0.665 & 15.803 & 0.699 & 15.117 & 0.678 & 15.863 & 0.693 & 16.469 & 0.704 & 15.548 & 0.688 & 2.415\\
LIME \cite{guo2016lime, guo2016lime2} & 11.154 & 0.591 & 11.828 & 0.610 & 11.517 & 0.607 & 12.638 & 0.628 & 13.613 & 0.653 & 12.150 & 0.618 & 2.432\\
HDRCNN \cite{HDRCNN} w/ RHT \cite{yang2018image}& 13.709 & 0.467 & 13.921 & 0.458 & 13.800 & 0.474 & 13.716 & 0.446 & 13.558 & 0.454 & 13.741 & 0.460 & 4.599\\
HDRCNN \cite{HDRCNN} w/ PS \cite{dayley2010photoshop}  & 15.812 & 0.667 & 16.970 & 0.699 & 16.428 & 0.681 & 17.301 & 0.687 & 18.650  & 0.702 & 17.032 & 0.687 & 2.267\\
DPED (iPhone) \cite{DPED}& 15.134 & 0.609 & 16.505 & 0.636 & 15.907 & 0.622 & 16.571 & 0.627 & 17.251 & 0.649 & 16.274 & 0.629 & 2.903\\
DPED (BlackBerry) \cite{DPED} & 16.910 & 0.642 & 18.649 & 0.713 & 17.606 & 0.653 & 18.070 & 0.679 & 18.217 & 0.668 & 17.890 & 0.671 & 2.564\\
DPED (Sony) \cite{DPED}& 17.419  & 0.675 & 18.636 & 0.701 & 18.020  & 0.683 &  17.554 & 0.660 & 17.778 & 0.663 & 17.881 & 0.676 & 2.806\\
DPE (HDR) \cite{DPE} & 15.690 & 0.614 & 16.548 & 0.626 & 16.305 & 0.626 & 16.147 & 0.615 & 16.341 & 0.633 & 16.206 & 0.623 & 2.417\\
DPE (U-FiveK) \cite{DPE} & 16.240 & 0.653 & 16.805 & 0.646 & 16.837 & 0.671 &  16.762 & 0.654 & 16.707 & 0.650 & 16.670 & 0.655 & 2.606 \\
DPE (S-FiveK) \cite{DPE} & 16.933 & 0.678 & 17.701 & 0.668 & 17.741 & 0.696 & 17.572 & 0.674 & 17.601 & 0.670 & 17.510 & 0.677 & 2.621\\
HQEC \cite{HQEC} $\textasteriskcentered$& 13.385 & 0.641 & 14.470 & 0.666 & 13.911 & 0.656 & 14.891 & 0.674 & 15.777 & 0.692 & 14.487 & 0.666 & 2.445\\
RetinexNet \cite{Chen2018Retinex}  & 10.759 & 0.585 & 11.613 & 0.596 & 11.135 & 0.605 & 11.987 & 0.615 & 12.671 & 0.636 & 11.633 & 0.607 & 3.105\\
Deep UPE \cite{DeepUPE} & 13.161 & 0.610 & 13.901 & 0.642 & 13.689 & 0.632 & 14.806 & 0.649 & 15.678 & 0.667 & 14.247 & 0.640 & 2.405\\ 
Zero-DCE \cite{guo2020zero}  & 11.643  & 0.536 & 12.555 & 0.539 & 12.058 & 0.544 & 12.964  & 0.548 & 13.769 & 0.580 & 12.598 & 0.549 & 2.865 \\
MSEC \cite{afifi2021learning} & 19.158 & 0.746 &  20.096 & 0.734 & 20.205 & 0.769 & 18.975 & 0.719 & 18.983 & 0.727 & 19.483 & 0.739 & \underline{2.251}\\
ENC \cite{huang2022exposure} & 20.170 & \underline{0.837} & 22.610 & \underline{0.896}
& 22.350 & 0.853 & \underline{20.880} & \underline{0.869} & \underline{21.060} & \underline{0.872} & \underline{21.410} & \underline{0.865} & \textbf{2.220}\\
FECNet \cite{huang2022deep} & \textbf{20.862} & 0.834 & \underline{22.761} & 0.867 & \underline{22.898} & \underline{0.875} & 20.386 & 0.846 & 19.597 & 0.837 & 21.301 & 0.852 & 2.397\\
IAT \cite{cui2022you} & 19.620 & 0.799 & 21.237 & 0.833 & 21.267 & 0.838 & 19.744 & 0.825 & 19.515 & 0.820 & 20.277 & 0.823 & 2.567\\
Ours & \underline{20.635} & \textbf{0.860} & \textbf{22.855} & \textbf{0.898} & \textbf{23.142} & \textbf{0.904} & \textbf{21.282} & \textbf{0.886} & \textbf{21.106} & \textbf{0.884} & \textbf{21.804} & \textbf{0.886} & 2.307\\
\hline

\end{tabular}
}
\end{center}
\vspace{-6mm}
\end{table*}

In order to verify the generalization ability of our method, we also conduct experiments on SICE dataset \cite{cai2018learning}. 
Compared with MSEC dataset, the SICE dataset is much smaller with the capacity of 1000 training, 24 validation and 60 testing image pairs, followed the partition by \cite{huang2022exposure}. Moreover, the under- and over-exposed scenes in SICE are more extreme. In implementation, the second and the last second subsets are used as underexposed and overexposed inputs, and the middle exposure subset is served as the ground truth.

In Table \ref{table:SICE}, our method achieves the highest PSNR and SSIM score in both under- and over-exposure scenes. For the combined testing set comprising underexposed and overexposed image pairs, our proposed method achieves a PSNR of 22.18dB and an SSIM of 0.7897.
The qualitative results on SICE datasets show that our method has a much superior ability to correct exposure under those extreme exposure conditions than previous methods.

\linespread{1.2}
\setlength{\tabcolsep}{1.4pt}
\begin{table}\footnotesize
\caption{Quantitative results on SICE \cite{cai2018learning} testing set in terms of PSNR/SSIM. \textbf{The best results are highlighted with bold.} \underline{The}  \underline{second-best results are highlighted with underline.}}
    \label{table:SICE}
    \centering
    \begin{tabular}{l p{1.8cm}<{\centering} p{1.8cm}<{\centering} p{1.8cm}<{\centering}}
    \toprule
    Method & Under & Over & Average \\ 
    \midrule
    CHANE \cite{reza2004realization} & 12.69/0.5037 & 10.21/0.4847 & 11.45/0.4942\\
    RetinexNet \cite{Chen2018Retinex} & 12.94/0.5171 & 12.87/0.5252 & 12.90/0.5212 \\
    Zero-DCE \cite{guo2020zero} & 16.92/0.6330 & 7.11/0.4292 & 12.02/0.5311 \\
    MSEC \cite{afifi2021learning} & 19.62/0.6512 & 17.59/0.6560 & 18.58/0.6536 \\
    ENC \cite{huang2022exposure} &  21.77/0.7052 & 19.57/\underline{0.7267} &  20.67/\underline{0.7160} \\
    FECNet \cite{huang2022deep} & \underline{22.01}/0.6737 & \underline{19.91}/0.6961 & \underline{20.96}/0.6849\\
    IAT \cite{cui2022you} & 20.73/\underline{0.7310} & 17.37/0.6630 & 19.05/0.6970\\ 
    Ours & \textbf{22.62}/\textbf{0.7818} & \textbf{21.73}/\textbf{0.7975} & \textbf{22.18}/\textbf{0.7897} \\ 
    \bottomrule
    \end{tabular}
\end{table}

\subsection{Qualitative Results}
\subsubsection{Visualization comparison with existing exposure correction methods.}
\begin{figure*}[htbp]
\centering
\includegraphics[width=0.8\linewidth]{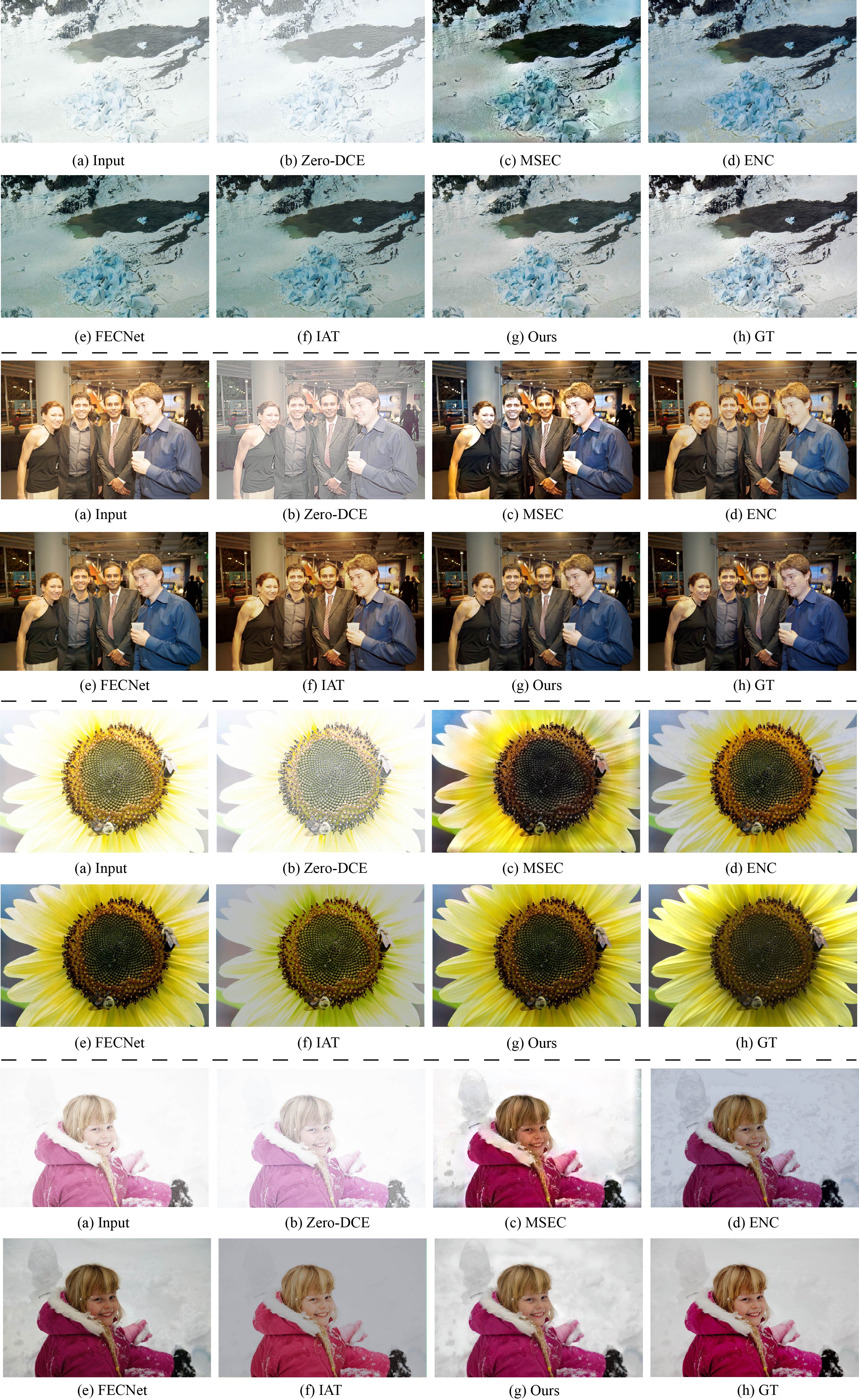}
\vspace{-2mm}
\caption{Visualization results on MSEC dataset \cite{afifi2021learning} of overexposure correction.\vspace{-6mm}}
\label{fig:MSEC_over}
\end{figure*}

\begin{figure*}[!htbp]
\centering
\includegraphics[width=0.8\linewidth]{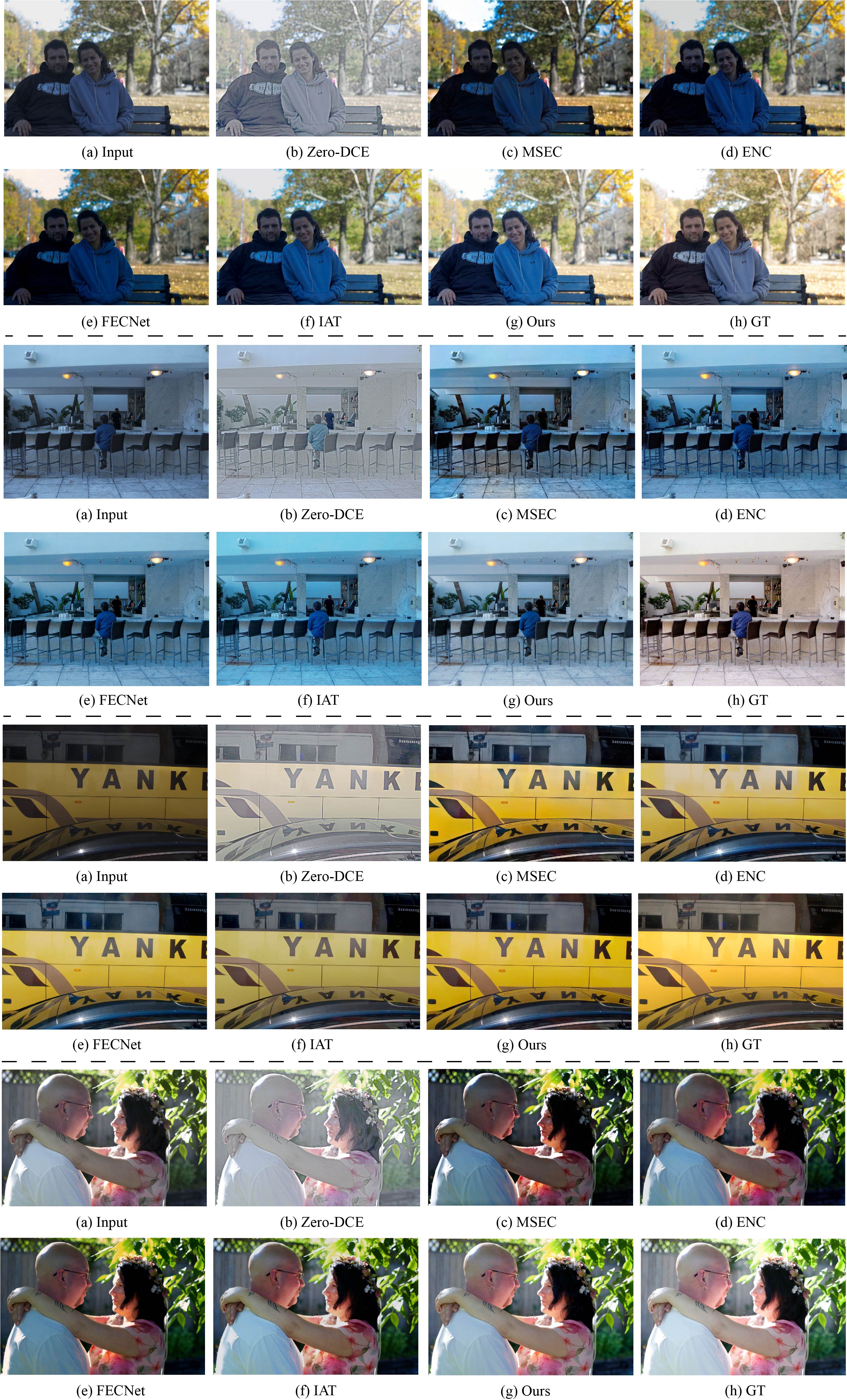}
\vspace{-2mm}
\caption{Visualization results on MSEC dataset \cite{afifi2021learning} of underexposure correction.\vspace{-2mm}}
\label{fig:MSEC_under}
\end{figure*}

We compare our methods with the five previous best performing models, Zero-DCE \cite{guo2020zero}, MSEC \cite{afifi2021learning}, ENC \cite{huang2022exposure}, FECNet \cite{huang2022deep} and IAT \cite{cui2022you}. Fig. \ref{fig:MSEC_over} and \ref{fig:MSEC_under} demonstrate the visualization results on MSEC dataset of over-exposure correction and under-exposure correction respectively. As can be seen in Fig. \ref{fig:MSEC_over}, as a low-light image enhancement method, Zero-DCE cannot effectively restore the details and colors of images for the task of exposure correction. For other exposure correction methods, they result in obvious color deviations and blocking artifacts.

For instance, as the first scene in Fig. \ref{fig:MSEC_over} shows, all the other previous methods fail to recover the true color of the snow, and the images they restore are all bluish. Comparatively, our method can restore the color of the image as close as possible to the original scene after the exposure correction procedure. There is a distinction between the brightest and the darkest in the generated image, as the lake in our generated image is clearer and more visible compared to the original over-exposed input.
The correction of overexposure at the edges for the overexposed sunflower in the third scene in Fig. \ref{fig:MSEC_over} is not favorable with both ENC and IAT.
Also, in the last scene in Fig. \ref{fig:MSEC_over}, MSEC and IAT fail to recover the background of the input image. Although FECNet can restore the background to be similar to ground truth, the little girl in the foreground is over-corrected to be darker than she should be. Our method can balance the relationship between the background and the foreground, making the overall image more harmonious and close to reality.

As for under-exposed scenarios in Fig. \ref{fig:MSEC_under}, all the previous methods cannot recover the facial features or the overall color as our method can do in the first two scenes. In the third scene in Fig. \ref{fig:MSEC_under}, some blocking artifacts appear around the letters on the yellow bus in the images generated by MSEC and ENC. The image generated by IAT suffers from color deviation. In contrast, our method can perform more visual friendly colors and spatial coherent contents, while introducing fewer artifacts as well. 

\begin{figure*}[h]
\centering
\includegraphics[width=0.92\linewidth]{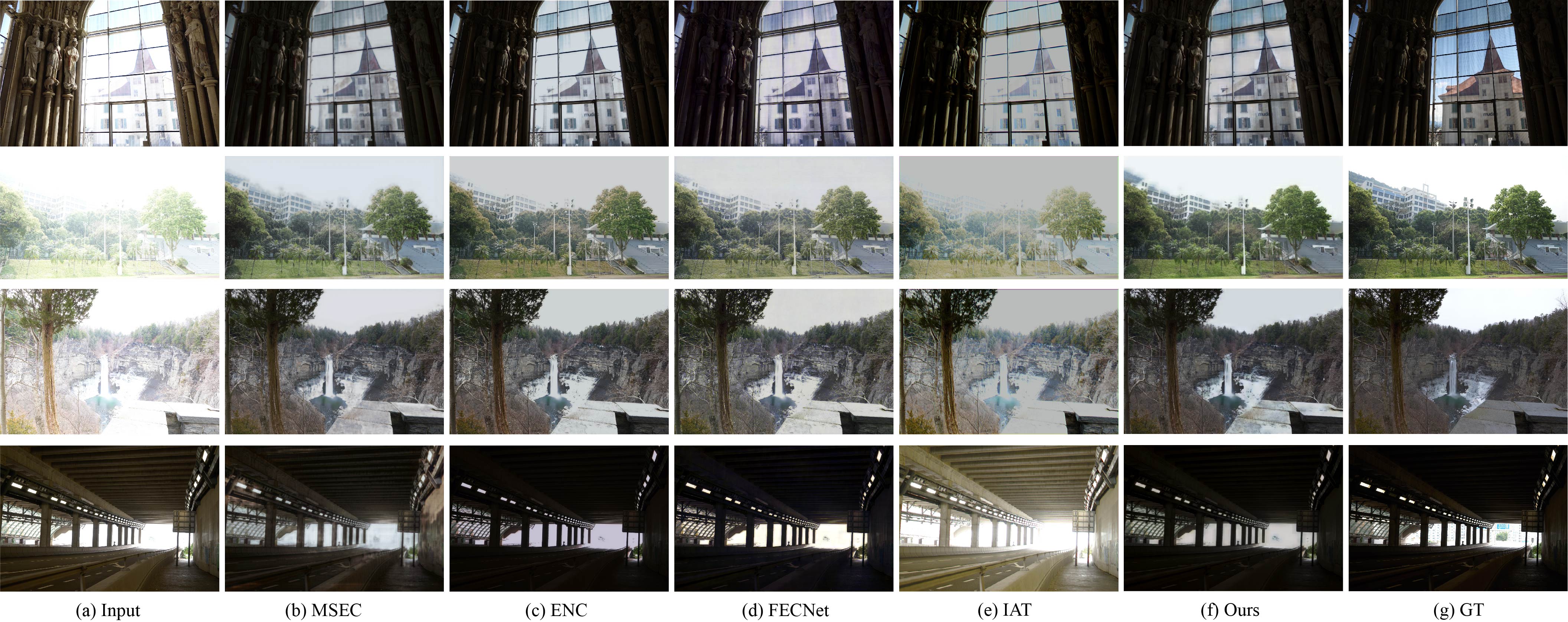}
\vspace{-3mm}
\caption{Visualization results on SICE dataset \cite{cai2018learning} of overexposure correction.\vspace{-4mm}}
\label{fig:SICE_over}
\end{figure*}

\begin{figure*}[h]
\centering
\includegraphics[width=0.92\linewidth]{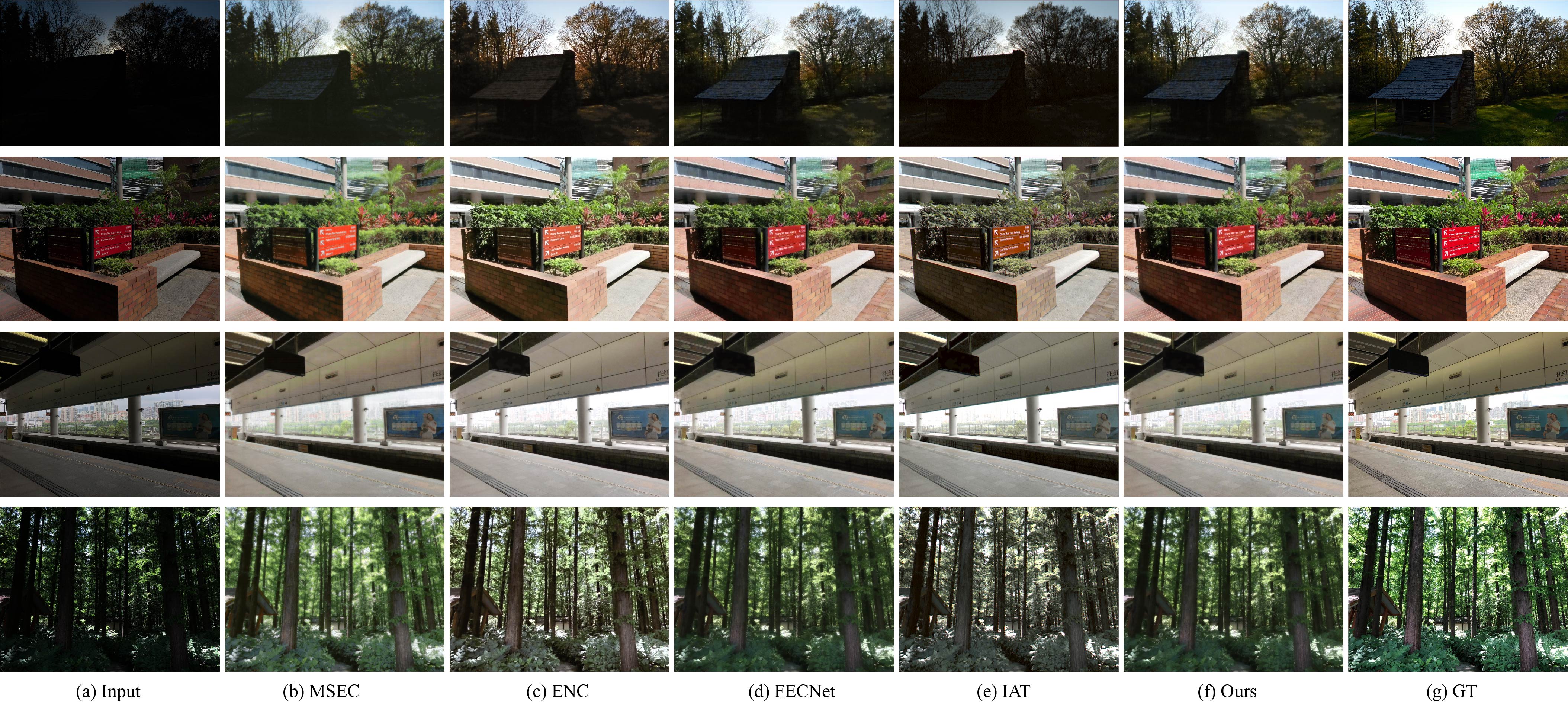}
\vspace{-3mm}
\caption{Visualization results on SICE dataset \cite{cai2018learning} of underexposure correction.\vspace{-4mm}}
\label{fig:SICE_under}
\end{figure*}

Apart from the experiments on MSEC datasets, in Fig. \ref{fig:SICE_over} and \ref{fig:SICE_under}, we also present visual comparison on the SICE dataset \cite{cai2018learning}. 
For the first row in Fig. \ref{fig:SICE_over}, previous methods fail to recover the architecture and the natural blue sky outside the window. Comparatively, we can observe that our proposed method can preserve more details and introduce less blocking artifacts under scenes with large lighting ratio. Similar observations can also be made in the following scenes. 

For under-exposed scenes, the difficulties of exposure correction lie in recovering the original color of the image and hallucinate the missing details in low-light conditions at the same time. Our method can recover the red brand and the green forest to the correctly exposed scene.

\subsubsection{Visualization comparison with commercial software packages.}
\begin{figure*}[htbp]
\centering
\includegraphics[width=0.93\linewidth]{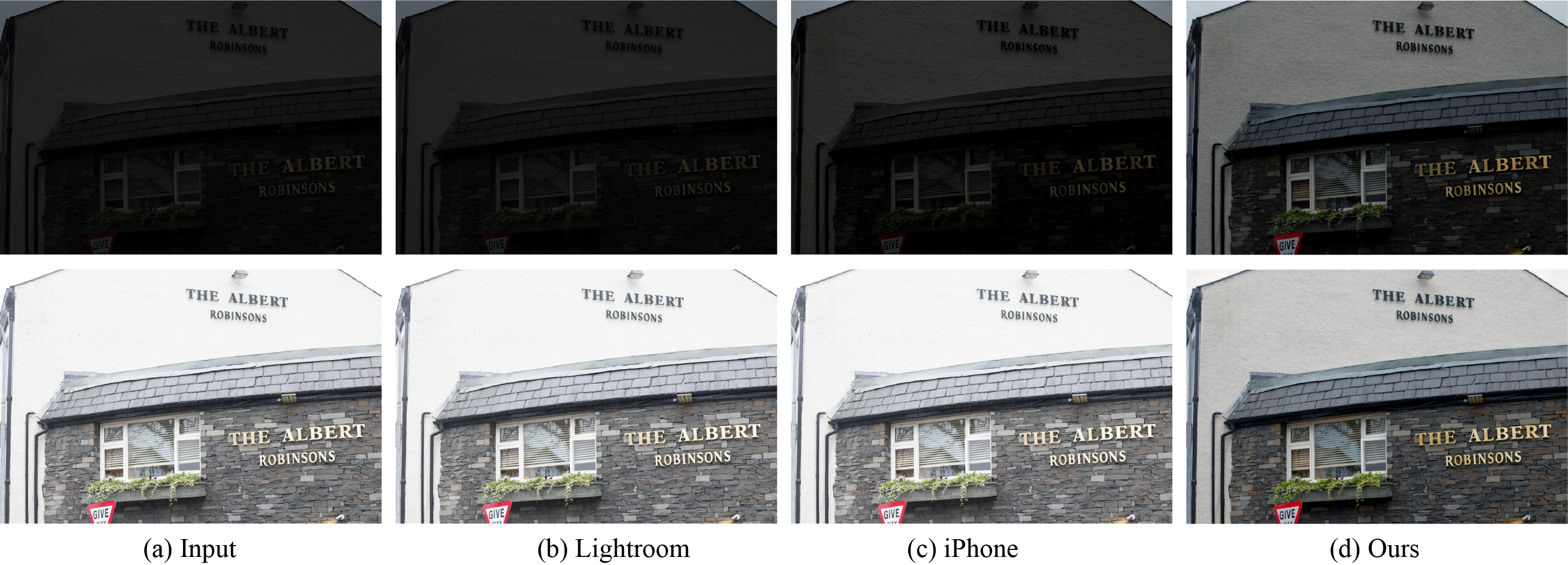}
\caption{Visualization comparison with commercial software packages on SICE dataset \vspace{-2mm}}
\label{fig:commercial_sice}
\end{figure*}

\begin{figure*}[htbp]
\centering
\includegraphics[width=0.93\linewidth]{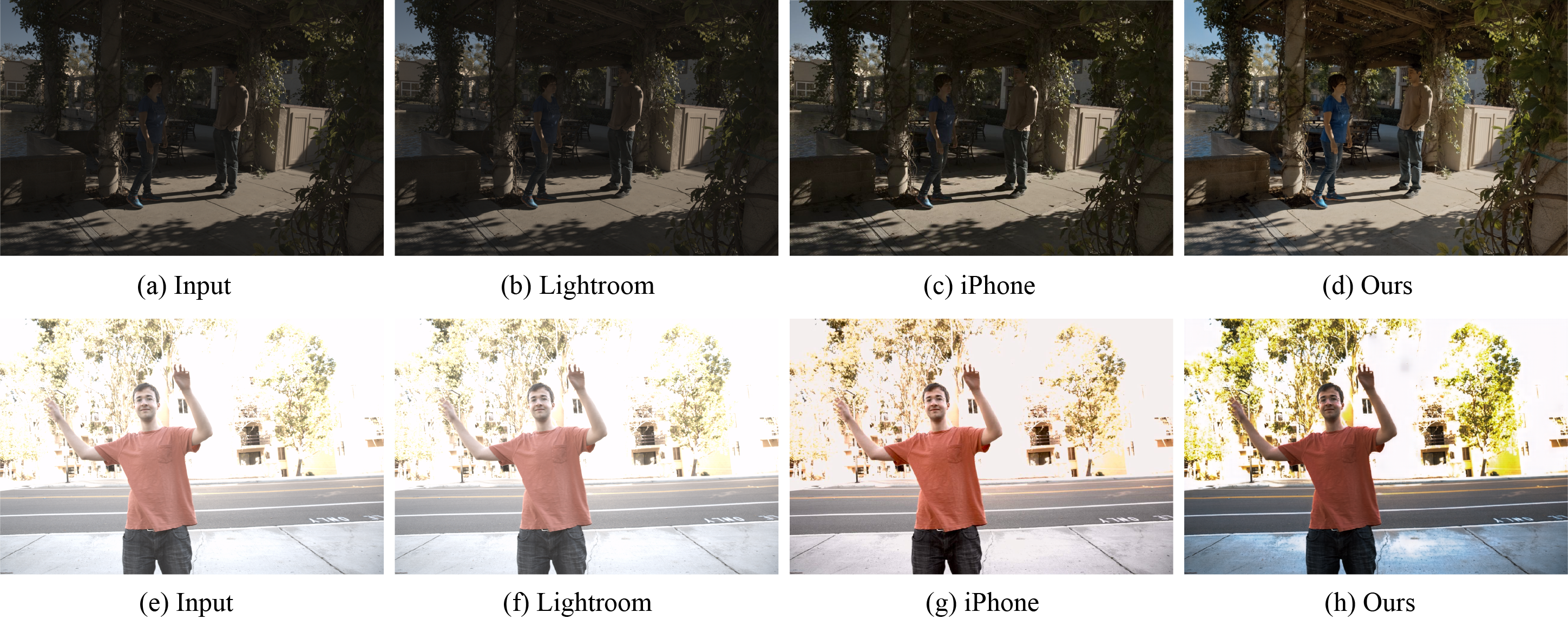}
\vspace{-2mm}
\caption{Visualization comparison with commercial software packages on HDR dataset \cite{kalantari2017deep} \vspace{-2mm}}
\label{fig:commercial_hdr}
\end{figure*}

Since there are already some off-the-shelf commercial software for image processing, which can also assist us in correcting exposure. In Fig. \ref{fig:commercial_sice} and \ref{fig:commercial_hdr}, we compare our proposed exposure correction method with the existing mainstream commercial photo editing software packages, Adobe Lightroom and iPhone Photo Enhancer, on SICE \cite{cai2018learning} and HDR \cite{kalantari2017deep} datasets, respecitvely. As for both Lightroom and iPhone, we apply auto enhancement on the testing images.

As can be seen from Fig. \ref{fig:commercial_sice}, the exposure adjustments in underexposed and overexposed scenes for both of the commercial software packages are relatively limited. Comparatively, our proposed exposure correction method is able to genuinely adjust the exposure of the image. 
For the underexposed and the overexposed images under the same scene Fig. \ref{fig:commercial_sice}(a), our exposure correction method as shown in Fig. \ref{fig:commercial_sice}(d) can correct the two images to a similar exposure level. While the underexposed and the overexposed images corrected by the other methods still have large differences in exposure level, testifying to the effectiveness of our proposed exposure correction method.

In Fig. \ref{fig:commercial_hdr}, the incorrectly exposed images are selected from the short- and long-exposed images in the HDR dataset with -2 EV and 2 EV deviations to the corresponding medium-exposed reference images. 
Adobe Lightroom does little modification on the image under auto enhancement mode as shown in Fig. \ref{fig:commercial_hdr}(b, f). 
iPhone Photo Enhancer can correct the overall exposure to some extent, which can be seen from the color change in the man's clothes in  Fig. \ref{fig:commercial_hdr}(g). But in Fig. \ref{fig:commercial_hdr}(c), people in the shadows are still so dark that it is hard to distinguish the facial expressions and clothing colors. In addition, the plants in the background cannot be recovered to a pleasing color in Fig. \ref{fig:commercial_hdr}(g).
Fig. \ref{fig:commercial_hdr}(d) reveals that our method is capable of enhancing the exposure of the image to an appropriate level. In particular, the colors of clothing in the shadow areas can be easily identified, and the details of the background vegetation behind the individuals are clear and distinguishable. Furthermore, the distant sky region does not become overexposed during the exposure enhancement. In the overexposed scene shown in Fig. \ref{fig:commercial_hdr}(h), although our method is unable to hallucinate all the details on the overexposed wall, the plants appearing in the background are restored to a green color that aligns with common perception. Regarding the foreground subjects, our method can restore them to a visually pleasing level without exhibiting a pale white appearance.

Moreover, our model is trained on the SICE dataset with limited number of training pairs compared to the mature commercial software packages, but it can still achieve satisfying visual experience.

\subsection{Ablation Study}
To explore the effectiveness of different parts in the architecture and losses of our method, we design the following ablation studies on the MSEC dataset. Table \ref{tab:ablation} summarizes the quantitative results of our ablation studies. The first row demonstrates the results when directly applying Transformer to exposure correction problem, failing to achieve a satisfactory performance. The following three rows probe into the effectiveness of different modules, namely SCAM, CAB and HINB. By applying SCAM in shallow feature extraction stage, features can be coarsely aligned and self-calibrated, leading to a performance gain of around 0.4dB. As for the CAB and HINB injected in Transformer blocks, they both improve PSNR by more than 1.9dB, compared with the results of Transformer-based method. The huge improvement verifies the effectiveness of our proposed CIT for exposure correction, for both CAB and HINB can restore local image details and alleviate the blocking artifacts. Performance can be further enhanced by injecting both of the two CNN modules, as shown in the last row of Table \ref{tab:ablation}. We also conduct experiments to verify the effectiveness of the losses by training our model from scratch. We can see that the color loss and spatial loss can bring a 0.1dB improvement to the performance of our proposed network.

\begin{figure}[h]
\centering
\includegraphics[width=\linewidth]{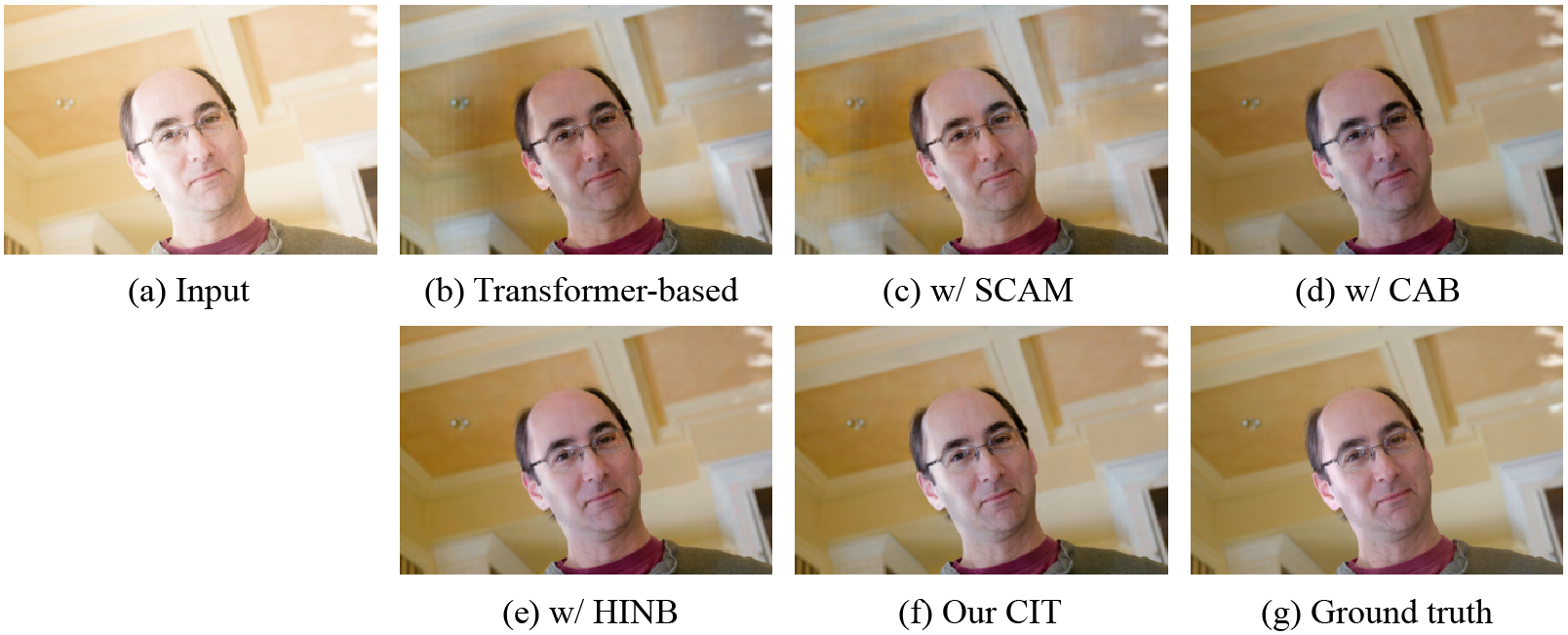}
\vspace{-8mm}
\caption{Ablation study on the modules in our CNN Injected Transformer.\vspace{-4mm}}
\label{fig:ablation}
\end{figure}

Apart from the numerical comparison, Fig. \ref{fig:ablation} visually demonstrates the effectiveness of the various modules in our proposed structure for addressing exposure correction.
We consider a window-based Transformer model as the baseline model, as demonstrated in Fig. \ref{fig:ablation}(b).
By utilizing Transformer's capability of constructing long-range feature dependencies, a window-based Transformer model can greatly rectify the exposure deviations. Nonetheless, blocking artifacts may be observed in the background regions.
With the aid of SCAM, the overall exposure level gets closer to that of the ground truth as the features are coarsely aligned, as depicted in Fig. \ref{fig:ablation}(c).
We incorporate Transformer blocks into both CAB and HINB to improve the visual representation.
In Fig. \ref{fig:ablation}(d, e), the results demonstrate that adding only CAB or HINB to Transformer can alleviate the blocking artifacts. However, the model performs better in global tone when both of them are added, as illustrated in Fig. \ref{fig:ablation}(f).

\setlength{\tabcolsep}{1.4pt}
\begin{table}\footnotesize
\vspace{-4mm}
\caption{Ablation study for investigating components of our method. (Testing on MSEC dataset)}
\label{tab:ablation}    
    \centering
    \setlength\tabcolsep{3pt}
    \begin{tabular}{cccccccc}
    \toprule
    SCAM & CAB & HINB & $L_1$ & $L_{col}$ & $L_{spa}$ & PSNR & SSIM \\ 
    \midrule
     &  &  & \checkmark & \checkmark & \checkmark & 19.766 & 0.875 \\
    \checkmark &  &  & \checkmark & \checkmark & \checkmark & 20.195 & 0.875 \\
    \checkmark & \checkmark &  & \checkmark & \checkmark & \checkmark & 21.676 & 0.884 \\
    \checkmark &  & \checkmark & \checkmark & \checkmark & \checkmark & 21.691 & 0.885 \\
    \checkmark & \checkmark & \checkmark & \checkmark &  &  &  21.655 &  0.884 \\
    \checkmark & \checkmark & \checkmark & \checkmark & \checkmark & \checkmark & \textbf{21.804} & \textbf{0.886} \\ 
    \bottomrule
    \end{tabular}
    \vspace{-6mm}
\end{table}
\setlength{\tabcolsep}{1.4pt}

\section{Conclusion}
\label{sec:con}
In this paper, we propose a CNN injected Transformer for exposure correction. In the low-resolution feature extraction stage, we introduce a self-calibrated attention module to model non-linear transformations between channels of feature maps and preserve the original image details. Furthermore, we couple CNN and Transformer to better aggregate short- and long-range information. We inject two convolutional-based module, namely channel attention block and half-instance normalization block, into the window-based Transformer block to enhance the representation ability of the network and eliminate possible artifacts.
Extensive experiments are conducted, validating the effectiveness of our model. Our method achieves state-of-the-art performance in both qualitative and quantitative metrics.

\bibliographystyle{IEEEtran}
\bibliography{CITEC}

\begin{thebibliography}{10}
\providecommand{\url}[1]{#1}
\csname url@samestyle\endcsname
\providecommand{\newblock}{\relax}
\providecommand{\bibinfo}[2]{#2}
\providecommand{\BIBentrySTDinterwordspacing}{\spaceskip=0pt\relax}
\providecommand{\BIBentryALTinterwordstretchfactor}{4}
\providecommand{\BIBentryALTinterwordspacing}{\spaceskip=\fontdimen2\font plus
\BIBentryALTinterwordstretchfactor\fontdimen3\font minus
  \fontdimen4\font\relax}
\providecommand{\BIBforeignlanguage}[2]{{%
\expandafter\ifx\csname l@#1\endcsname\relax
\typeout{** WARNING: IEEEtran.bst: No hyphenation pattern has been}%
\typeout{** loaded for the language `#1'. Using the pattern for}%
\typeout{** the default language instead.}%
\else
\language=\csname l@#1\endcsname
\fi
#2}}
\providecommand{\BIBdecl}{\relax}
\BIBdecl

\bibitem{afifi2021learning}
M.~Afifi, K.~G. Derpanis, B.~Ommer, and M.~S. Brown, ``Learning multi-scale
  photo exposure correction,'' in \emph{IEEE Conf. Comput. Vis. Pattern
  Recog.}, 2021, pp. 9157--9167.

\bibitem{liang2021swinir}
J.~Liang, J.~Cao, G.~Sun, K.~Zhang, L.~Van~Gool, and R.~Timofte, ``Swinir:
  Image restoration using swin transformer,'' in \emph{IEEE Conf. Comput. Vis.
  Pattern Recog.}, 2021, pp. 1833--1844.

\bibitem{cui2022you}
Z.~Cui, K.~Li, L.~Gu, S.~Su, P.~Gao, Z.~Jiang, Y.~Qiao, and T.~Harada, ``You
  only need 90k parameters to adapt light: A light weight transformer for image
  enhancement and exposure correction,'' in \emph{Brit. Mach. Vis. Conf.},
  2022, pp. 21--24.

\bibitem{wang2019underexposed}
R.~Wang, Q.~Zhang, C.-W. Fu, X.~Shen, W.-S. Zheng, and J.~Jia, ``Underexposed
  photo enhancement using deep illumination estimation,'' in \emph{IEEE Conf.
  Comput. Vis. Pattern Recog.}, 2019, pp. 6849--6857.

\bibitem{srinivas2019exposure}
K.~Srinivas, A.~K. Bhandari, and A.~Singh, ``Exposure-based energy curve
  equalization for enhancement of contrast distorted images,'' \emph{IEEE
  Trans. Circuit Syst. Video Technol.}, vol.~30, no.~12, pp. 4663--4675, 2019.

\bibitem{zheng2020single}
C.~Zheng, Z.~Li, Y.~Yang, and S.~Wu, ``Single image brightening via multi-scale
  exposure fusion with hybrid learning,'' \emph{IEEE Trans. Circuit Syst. Video
  Technol.}, vol.~31, no.~4, pp. 1425--1435, 2020.

\bibitem{liu2022efinet}
C.~Liu, F.~Wu, and X.~Wang, ``Efinet: Restoration for low-light images via
  enhancement-fusion iterative network,'' \emph{IEEE Trans. Circuit Syst. Video
  Technol.}, vol.~32, no.~12, pp. 8486--8499, 2022.

\bibitem{li2021low}
J.~Li, X.~Feng, and Z.~Hua, ``Low-light image enhancement via
  progressive-recursive network,'' \emph{IEEE Trans. Circuit Syst. Video
  Technol.}, vol.~31, no.~11, pp. 4227--4240, 2021.

\bibitem{fan2022multiscale}
G.-D. Fan, B.~Fan, M.~Gan, G.-Y. Chen, and C.~P. Chen, ``Multiscale low-light
  image enhancement network with illumination constraint,'' \emph{IEEE Trans.
  Circuit Syst. Video Technol.}, vol.~32, no.~11, pp. 7403--7417, 2022.

\bibitem{huang2022exposure}
J.~Huang, Y.~Liu, X.~Fu, M.~Zhou, Y.~Wang, F.~Zhao, and Z.~Xiong, ``Exposure
  normalization and compensation for multiple-exposure correction,'' in
  \emph{IEEE Conf. Comput. Vis. Pattern Recog.}, 2022, pp. 6043--6052.

\bibitem{huang2022deep}
J.~Huang, Y.~Liu, F.~Zhao, K.~Yan, J.~Zhang, Y.~Huang, M.~Zhou, and Z.~Xiong,
  ``Deep fourier-based exposure correction network with spatial-frequency
  interaction,'' in \emph{Eur. Conf. Comput. Vis.}\hskip 1em plus 0.5em minus
  0.4em\relax Springer, 2022, pp. 163--180.

\bibitem{vaswani2017attention}
A.~Vaswani, N.~Shazeer, N.~Parmar, J.~Uszkoreit, L.~Jones, A.~N. Gomez,
  {\L}.~Kaiser, and I.~Polosukhin, ``Attention is all you need,'' \emph{Adv.
  Neural Inform. Process. Syst.}, vol.~30, 2017.

\bibitem{dosovitskiy2020image}
A.~Dosovitskiy, L.~Beyer, A.~Kolesnikov, D.~Weissenborn, X.~Zhai,
  T.~Unterthiner, M.~Dehghani, M.~Minderer, G.~Heigold, S.~Gelly \emph{et~al.},
  ``An image is worth 16x16 words: Transformers for image recognition at
  scale,'' \emph{arXiv preprint arXiv:2010.11929}, 2020.

\bibitem{li2021two}
J.~Li, D.~Wang, X.~Liu, Z.~Shi, and M.~Wang, ``Two-branch attention network via
  efficient semantic coupling for one-shot learning,'' \emph{IEEE Trans. Image
  Process.}, vol.~31, pp. 341--351, 2021.

\bibitem{celik2011contextual}
T.~Celik and T.~Tjahjadi, ``Contextual and variational contrast enhancement,''
  \emph{IEEE Trans. Image Process.}, vol.~20, no.~12, pp. 3431--3441, 2011.

\bibitem{lee2013contrast}
C.~Lee, C.~Lee, and C.-S. Kim, ``Contrast enhancement based on layered
  difference representation of 2d histograms,'' \emph{IEEE Trans. Image
  Process.}, vol.~22, no.~12, pp. 5372--5384, 2013.

\bibitem{reza2004realization}
A.~M. Reza, ``Realization of the contrast limited adaptive histogram
  equalization (clahe) for real-time image enhancement,'' \emph{J. Signal
  Process. Syst}, vol.~38, no.~1, pp. 35--44, 2004.

\bibitem{thomas2011histogram}
G.~Thomas, D.~Flores-Tapia, and S.~Pistorius, ``Histogram specification: a fast
  and flexible method to process digital images,'' \emph{IEEE Transactions on
  Instrumentation and Measurement}, vol.~60, no.~5, pp. 1565--1578, 2011.

\bibitem{guo2016lime}
X.~Guo, ``Lime: A method for low-light image enhancement,'' in \emph{ACM Int.
  Conf. Multimedia}, 2016, pp. 87--91.

\bibitem{zhang2018high}
Q.~Zhang, G.~Yuan, C.~Xiao, L.~Zhu, and W.-S. Zheng, ``High-quality exposure
  correction of underexposed photos,'' in \emph{ACM Int. Conf. Multimedia},
  2018, pp. 582--590.

\bibitem{meylan2006high}
L.~Meylan and S.~Susstrunk, ``High dynamic range image rendering with a
  retinex-based adaptive filter,'' \emph{IEEE Trans. Image Process.}, vol.~15,
  no.~9, pp. 2820--2830, 2006.

\bibitem{cai2017joint}
B.~Cai, X.~Xu, K.~Guo, K.~Jia, B.~Hu, and D.~Tao, ``A joint intrinsic-extrinsic
  prior model for retinex,'' in \emph{IEEE Conf. Comput. Vis. Pattern Recog.},
  2017, pp. 4000--4009.

\bibitem{guo2016lime2}
X.~Guo, Y.~Li, and H.~Ling, ``Lime: Low-light image enhancement via
  illumination map estimation,'' \emph{IEEE Trans. Image Process.}, vol.~26,
  no.~2, pp. 982--993, 2016.

\bibitem{fu2016fusion}
X.~Fu, D.~Zeng, Y.~Huang, Y.~Liao, X.~Ding, and J.~Paisley, ``A fusion-based
  enhancing method for weakly illuminated images,'' \emph{Signal Process.},
  vol. 129, pp. 82--96, 2016.

\bibitem{nsamp2021learning}
N.~E. Nsampi, Z.~Hu, and Q.~Wang, ``Learning exposure correction via
  consistency modeling,'' in \emph{Proc. Brit. Mach. Vision Conf.}, 2021.

\bibitem{huang2019hybrid}
J.~Huang, Z.~Xiong, X.~Fu, D.~Liu, and Z.-J. Zha, ``Hybrid image enhancement
  with progressive laplacian enhancing unit,'' in \emph{ACM Int. Conf.
  Multimedia}, 2019, pp. 1614--1622.

\bibitem{jiang2021enlightengan}
Y.~Jiang, X.~Gong, D.~Liu, Y.~Cheng, C.~Fang, X.~Shen, J.~Yang, P.~Zhou, and
  Z.~Wang, ``Enlightengan: Deep light enhancement without paired supervision,''
  \emph{IEEE Trans. Image Process.}, vol.~30, pp. 2340--2349, 2021.

\bibitem{liu2021retinex}
R.~Liu, L.~Ma, J.~Zhang, X.~Fan, and Z.~Luo, ``Retinex-inspired unrolling with
  cooperative prior architecture search for low-light image enhancement,'' in
  \emph{IEEE Conf. Comput. Vis. Pattern Recog.}, 2021, pp. 10\,561--10\,570.

\bibitem{huang2022low}
J.~Huang, X.~Fu, Z.~Xiao, F.~Zhao, and Z.~Xiong, ``Low-light stereo image
  enhancement,'' \emph{IEEE Trans. Multimedia}, 2022.

\bibitem{eyiokur2022exposure}
F.~Eyiokur, D.~Yaman, H.~K. Ekenel, and A.~Waibel, ``Exposure correction model
  to enhance image quality,'' in \emph{IEEE Conf. Comput. Vis. Pattern Recog.},
  2022.

\bibitem{wang2023decoupling}
Y.~Wang, L.~Peng, L.~Li, Y.~Cao, and Z.-J. Zha, ``Decoupling-and-aggregating
  for image exposure correction,'' in \emph{Proceedings of the IEEE/CVF
  Conference on Computer Vision and Pattern Recognition}, 2023, pp.
  18\,115--18\,124.

\bibitem{huang2023learning}
J.~Huang, F.~Zhao, M.~Zhou, J.~Xiao, N.~Zheng, K.~Zheng, and Z.~Xiong,
  ``Learning sample relationship for exposure correction,'' in
  \emph{Proceedings of the IEEE/CVF Conference on Computer Vision and Pattern
  Recognition}, 2023, pp. 9904--9913.

\bibitem{vaswani2021scaling}
A.~Vaswani, P.~Ramachandran, A.~Srinivas, N.~Parmar, B.~Hechtman, and
  J.~Shlens, ``Scaling local self-attention for parameter efficient visual
  backbones,'' in \emph{IEEE Conf. Comput. Vis. Pattern Recog.}, 2021, pp.
  12\,894--12\,904.

\bibitem{carion2020end}
N.~Carion, F.~Massa, G.~Synnaeve, N.~Usunier, A.~Kirillov, and S.~Zagoruyko,
  ``End-to-end object detection with transformers,'' in \emph{Eur. Conf.
  Comput. Vis.}\hskip 1em plus 0.5em minus 0.4em\relax Springer, 2020, pp.
  213--229.

\bibitem{chu2021twins}
X.~Chu, Z.~Tian, Y.~Wang, B.~Zhang, H.~Ren, X.~Wei, H.~Xia, and C.~Shen,
  ``Twins: Revisiting the design of spatial attention in vision transformers,''
  \emph{Adv. Neural Inform. Process. Syst.}, pp. 9355--9366, 2021.

\bibitem{wang2021pyramid}
W.~Wang, E.~Xie, X.~Li, D.-P. Fan, K.~Song, D.~Liang, T.~Lu, P.~Luo, and
  L.~Shao, ``Pyramid vision transformer: A versatile backbone for dense
  prediction without convolutions,'' in \emph{IEEE Conf. Comput. Vis. Pattern
  Recog.}, 2021, pp. 568--578.

\bibitem{chen2021pre}
H.~Chen, Y.~Wang, T.~Guo, C.~Xu, Y.~Deng, Z.~Liu, S.~Ma, C.~Xu, C.~Xu, and
  W.~Gao, ``Pre-trained image processing transformer,'' in \emph{IEEE Conf.
  Comput. Vis. Pattern Recog.}, 2021, pp. 12\,299--12\,310.

\bibitem{cao2021video}
J.~Cao, Y.~Li, K.~Zhang, and L.~Van~Gool, ``Video super-resolution
  transformer,'' \emph{arXiv preprint arXiv:2106.06847}, 2021.

\bibitem{zamir2022restormer}
S.~W. Zamir, A.~Arora, S.~Khan, M.~Hayat, F.~S. Khan, and M.-H. Yang,
  ``Restormer: Efficient transformer for high-resolution image restoration,''
  in \emph{IEEE Conf. Comput. Vis. Pattern Recog.}, 2022, pp. 5728--5739.

\bibitem{chen2023activating}
X.~Chen, X.~Wang, J.~Zhou, Y.~Qiao, and C.~Dong, ``Activating more pixels in
  image super-resolution transformer,'' in \emph{IEEE Conf. Comput. Vis.
  Pattern Recog.}, 2023, pp. 22\,367--22\,377.

\bibitem{wang2022uformer}
Z.~Wang, X.~Cun, J.~Bao, W.~Zhou, J.~Liu, and H.~Li, ``Uformer: A general
  u-shaped transformer for image restoration,'' in \emph{IEEE Conf. Comput.
  Vis. Pattern Recog.}, 2022, pp. 17\,683--17\,693.

\bibitem{liu2021swin}
Z.~Liu, Y.~Lin, Y.~Cao, H.~Hu, Y.~Wei, Z.~Zhang, S.~Lin, and B.~Guo, ``Swin
  transformer: Hierarchical vision transformer using shifted windows,'' in
  \emph{IEEE Conf. Comput. Vis. Pattern Recog.}, 2021, pp. 10\,012--10\,022.

\bibitem{song2022multistage}
B.~Song, J.~Zhou, and H.~Wu, ``Multistage curvature-guided network for
  progressive single image reflection removal,'' \emph{IEEE Transactions on
  Circuits and Systems for Video Technology}, vol.~32, no.~10, pp. 6515--6529,
  2022.

\bibitem{song2023real}
B.~Song, J.~Zhou, X.~Chen, and S.~Zhang, ``Real-scene reflection removal with
  raw-rgb image pairs,'' \emph{IEEE Transactions on Circuits and Systems for
  Video Technology}, 2023.

\bibitem{chen2021hinet}
L.~Chen, X.~Lu, J.~Zhang, X.~Chu, and C.~Chen, ``Hinet: Half instance
  normalization network for image restoration,'' in \emph{IEEE Conf. Comput.
  Vis. Pattern Recog.}, 2021, pp. 182--192.

\bibitem{guo2020zero}
C.~Guo, C.~Li, J.~Guo, C.~C. Loy, J.~Hou, S.~Kwong, and R.~Cong,
  ``Zero-reference deep curve estimation for low-light image enhancement,'' in
  \emph{IEEE Conf. Comput. Vis. Pattern Recog.}, 2020, pp. 1780--1789.

\bibitem{buchsbaum1980spatial}
G.~Buchsbaum, ``A spatial processor model for object colour perception,''
  \emph{J Franklin Inst}, vol. 310, no.~1, pp. 1--26, 1980.

\bibitem{bychkovsky2011learning}
V.~Bychkovsky, S.~Paris, E.~Chan, and F.~Durand, ``Learning photographic global
  tonal adjustment with a database of input/output image pairs,'' in \emph{IEEE
  Conf. Comput. Vis. Pattern Recog.}\hskip 1em plus 0.5em minus 0.4em\relax
  IEEE, 2011, pp. 97--104.

\bibitem{gonzales2001digital}
R.~C. Gonzales and R.~E. Woods, ``Digital image processing second edition,''
  2001.

\bibitem{adaptivehisteq}
K.~Zuiderveld, ``Contrast limited adaptive histogram equalization,''
  \emph{Graphics gems}, pp. 474--485, 1994.

\bibitem{fu2016weighted}
X.~Fu, D.~Zeng, Y.~Huang, X.-P. Zhang, and X.~Ding, ``A weighted variational
  model for simultaneous reflectance and illumination estimation,'' in
  \emph{IEEE Conf. Comput. Vis. Pattern Recog.}, 2016, pp. 2782--2790.

\bibitem{HDRCNN}
G.~Eilertsen, J.~Kronander, G.~Denes, R.~K. Mantiuk, and J.~Unger, ``Hdr image
  reconstruction from a single exposure using deep cnns,'' \emph{ACM Trans.
  Graph.}, vol.~36, no.~6, pp. 1--15, 2017.

\bibitem{DPED}
A.~Ignatov, N.~Kobyshev, R.~Timofte, K.~Vanhoey, and L.~Van~Gool,
  ``Dslr-quality photos on mobile devices with deep convolutional networks,''
  in \emph{Int. Conf. Comput. Vis.}, 2017, pp. 3277--3285.

\bibitem{DPE}
Y.-S. Chen, Y.-C. Wang, M.-H. Kao, and Y.-Y. Chuang, ``Deep photo enhancer:
  Unpaired learning for image enhancement from photographs with gans,'' in
  \emph{IEEE Conf. Comput. Vis. Pattern Recog.}, 2018, pp. 6306--6314.

\bibitem{HQEC}
Q.~Zhang, G.~Yuan, C.~Xiao, L.~Zhu, and W.-S. Zheng, ``High-quality exposure
  correction of underexposed photos,'' in \emph{ACM Int. Conf. Multimedia},
  2018, pp. 582--590.

\bibitem{Chen2018Retinex}
C.~Wei, W.~Wang, W.~Yang, and J.~Liu, ``Deep retinex decomposition for
  low-light enhancement,'' \emph{arXiv preprint arXiv:1808.04560}, 2018.

\bibitem{DeepUPE}
R.~Wang, Q.~Zhang, C.-W. Fu, X.~Shen, W.-S. Zheng, and J.~Jia, ``Underexposed
  photo enhancement using deep illumination estimation,'' in \emph{IEEE Conf.
  Comput. Vis. Pattern Recog.}, 2019, pp. 6849--6857.

\bibitem{wang2004image}
Z.~Wang, A.~C. Bovik, H.~R. Sheikh, and E.~P. Simoncelli, ``Image quality
  assessment: from error visibility to structural similarity,'' \emph{IEEE
  Trans. Image Process.}, vol.~13, no.~4, pp. 600--612, 2004.

\bibitem{blau20182018}
Y.~Blau, R.~Mechrez, R.~Timofte, T.~Michaeli, and L.~Zelnik-Manor, ``The 2018
  pirm challenge on perceptual image super-resolution,'' in \emph{Eur. Conf.
  Comput. Vis.}, 2018, pp. 0--0.

\bibitem{ma2017learning}
C.~Ma, C.-Y. Yang, X.~Yang, and M.-H. Yang, ``Learning a no-reference quality
  metric for single-image super-resolution,'' \emph{Comput Vis Image Underst},
  vol. 158, pp. 1--16, 2017.

\bibitem{mittal2012making}
A.~Mittal, R.~Soundararajan, and A.~C. Bovik, ``Making a “completely blind”
  image quality analyzer,'' \emph{IEEE Sign. Process. Letters}, vol.~20, no.~3,
  pp. 209--212, 2012.

\bibitem{yang2018image}
X.~Yang, K.~Xu, Y.~Song, Q.~Zhang, X.~Wei, and R.~W. Lau, ``Image correction
  via deep reciprocating hdr transformation,'' in \emph{IEEE Conf. Comput. Vis.
  Pattern Recog.}, 2018, pp. 1798--1807.

\bibitem{dayley2010photoshop}
L.~D. Dayley and B.~Dayley, \emph{Photoshop CS5 Bible}.\hskip 1em plus 0.5em
  minus 0.4em\relax John Wiley \& Sons, 2010.

\bibitem{cai2018learning}
J.~Cai, S.~Gu, and L.~Zhang, ``Learning a deep single image contrast enhancer
  from multi-exposure images,'' \emph{IEEE Trans. Image Process.}, vol.~27,
  no.~4, pp. 2049--2062, 2018.

\bibitem{kalantari2017deep}
N.~K. Kalantari, R.~Ramamoorthi \emph{et~al.}, ``Deep high dynamic range
  imaging of dynamic scenes.'' \emph{ACM Trans. Graph.}, vol.~36, no.~4, pp.
  144--1, 2017.

\end{thebibliography}

\end{document}